\documentclass[journal]{IEEEtran}

\usepackage{pdfpages}
\usepackage{amssymb,amsmath}
\usepackage{bm} 
\usepackage{verbatim} 
\usepackage{graphicx}
\usepackage{mysubfigure}
\usepackage{multirow}
\usepackage[font=small,labelfont=bf,margin=\parindent,tableposition=top]{caption}
\usepackage{algorithm}
\usepackage{algorithmic} 
\usepackage{arydshln}
\usepackage{framed}
\usepackage{cite}
\usepackage{times}
\usepackage{graphicx}
\usepackage{epstopdf}
\usepackage{hyperref} 
\usepackage{booktabs}
\usepackage{caption}
\usepackage{bbm}
\usepackage{dsfont}
\usepackage{verbatim}
\usepackage{epstopdf}

\newcommand{\myverb}{\url{http://www.sfu.ca/~amadooei/research/publication/TMI2015\_sup.html}}

\begin{document}

\title{Learning to Detect Blue-white Structures in Dermoscopy Images with Weak Supervision}

\author{Ali~Madooei,~
\thanks{The authors are with the Vision and Media Lab, 
School of Computing Science, 
Simon Fraser University, Canada. \textit{Emails: $\{$amadooei,~mark,~hosseinh$\}$@cs.sfu.ca}}
~~Mark~S.~Drew,~
~~Hossein~Hajimirsadeghi}

%



\maketitle

\begin{abstract}
We propose a novel approach to identify one of the most significant dermoscopic
criteria in the diagnosis of Cutaneous Melanoma: the Blue-whitish structure. In
this paper, we achieve this goal in a Multiple Instance Learning framework using
only image-level labels of whether the feature is present or not. As the output,
we predict the image classification label and as well localize the feature in the image. Experiments are conducted on a challenging dataset with results outperforming state-of-the-art. This study provides an improvement on the scope of modelling for computerized image analysis of skin lesions, in particular in that it puts forward a framework for identification of dermoscopic local features from weakly-labelled data.
\end{abstract}


%
\IEEEpeerreviewmaketitle

\section{Introduction}
\IEEEPARstart{D}{ermatological} practice relies heavily on visual examination of skin lesions.  
Thus, it is not surprising that interest in Computer Vision based diagnosis
technology in this field is growing. 
The goal of automatically understanding dermatological images is tremendously
challenging, however, since much like human vision itself
what is understood about how diagnostic expertise actually operates is subjective and limited.

Many of those who take up this challenge focus on detection of Cutaneous (skin)
Melanoma through dermoscopy image analysis. Melanoma is the most
life-threatening form of skin cancer. Dermoscopy is a non-invasive, in-vivo skin examination technique that uses optical magnification and cross-polarized lighting to allow enhanced visualization of lesion characteristics which are often not discernible by the naked eye. 
Early detection of melanoma is paramount to patients' prognosis
towards greater survival. The challenges involved in clinical diagnosis of early melanoma have
provoked increased interest in computer-aided diagnosis 
systems through automatic analysis of digital dermoscopy images.

In this paper, we focus on the identification of blue-whitish structures (BWS),
one of the most important findings through dermoscopic examination in making a
diagnosis of invasive melanoma \cite{soyer_three-point_2004}. The term Blue-white
structures is a unified heading for features also known as Blue-white veil and
Regression structures (this is discussed below in \S\ref{clinic}).

To this aim, a typical approach would be based on the classical paradigm of
supervised learning requiring extensively annotating each dermoscopic image with instances of BWS, in all training images. 
This is difficult (or even impossible) to be carried accurately and consistently due to subjectivity of feature definition and poor inter-observer agreement.

The dermoscopy data in fact available to us has motivated a different, more challenging, research problem. 
In this dataset \cite{argenziano_interactive_2000}, image-level labels encode
only whether an image contains a dermoscopic feature or not, but the features
themselves are not locally annotated.
In Computer Vision this situation is referred to as weakly-labelled data. 

To approach this problem, we use the multiple instance learning (MIL) paradigm.
MIL is a relatively new learning paradigm, and has broad applications in computer vision, text processing, etc. Unlike standard supervised learning, where each training instance is labeled, MIL is a type of weakly supervised learning, where the instance labels are ambiguous (more on this in \S\ref{mil}). 

Our goal is to learn to identify and localize BWS using this weakly-labelled data (i.e.\ with minimal supervision).
Learning to localize dermoscopic features with minimal supervision is an
important class of problem. It provides an improvement on the scope of modelling
for computerized image analysis of skin lesions because the vast majority of
data available are in fact weakly-labeled. Surprisingly, this class of problem
is the least studied in the relevant literature.

\section{Clinical background} \label{clinic}
Dermoscopy allows the identification of many different structures not seen by
the unaided eye. As the field has evolved, terminology has been accumulated to
describe structures seen via dermoscopy. This terminology can sometimes be
confusing. For clarity, a brief description follows to illuminate the feature under study here.

In dermoscopic examination of pigmented skin lesions, accurate analysis of lesion colouration is essential to the diagnosis. Lesions with dark, bluish or variegated colours are deemed to be more likely to be malignant. Indeed, the crucial role of colour cues is evident as most clinical diagnosis guidelines (such as the ``ABCD rule'' \cite{stolz_abcd_1994} and the ``7-point checklist'' \cite{argenzianog_epiluminescence_1998}) include colour for lesion scoring. 

Among the common colours seen under dermoscopy, the presence of a blue hue and a white (together or separately) are a diagnosis clue. White is often the result of depigmentation, sclerosis of the dermis or keratinization. 
Blue itself is the result of the Tyndall effect: the longer-wavelength light (red) is more transmitted while the shorter-wavelength light (blue) is more reflected via scattering from the melanin pigment present deep within the lesion.

Identification of blue and white hues within a lesion is a good predictor of
malignancy but not a specific one. Shades of blue are also observed in benign
lesions such as in blue nevi (moles) and haemangiomas, and white areas are also
seen in e.g.\ benign halo nevi. As a general rule, colours in melanoma are
focal, asymmetrical and irregular whereas in benign lesions are distributed
uniformly. This generalization however lacks adequate specificity. To overcome
these issues, two specific features have been defined, denoted as blue-white
veil feature and regression structure.

\begin{figure}[btp]
\centering
\includegraphics[width=0.22 \textwidth]{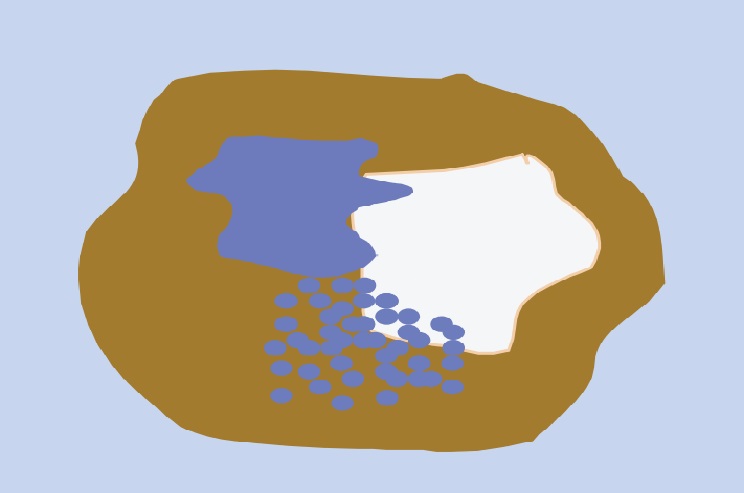}~~~
\includegraphics[width=0.22 \textwidth]{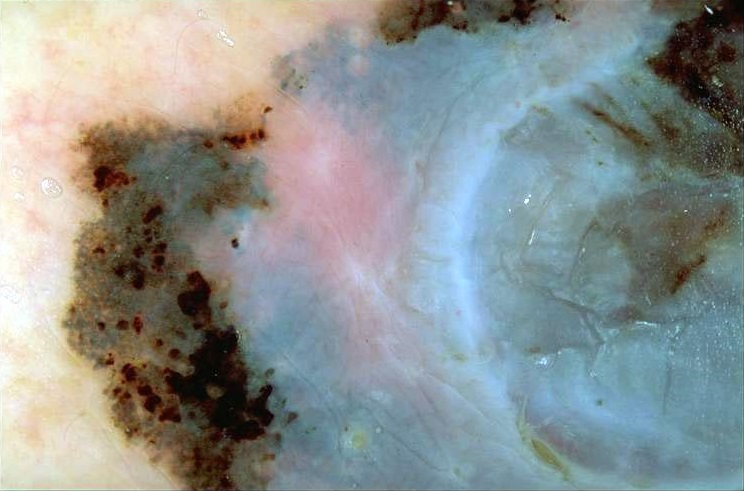}
\caption{\small{Schematic representation of blue-whitish structures (left) and a
dermoscopic image (right) of melanoma with this feature (It is difficult to differentiate
between `regression' and `blue-white veil' areas; both features are present).}}
\label{FIG:BWS}
\end{figure}

Blue-white veil is defined as irregular, confluent, grey-blue to whitish blue diffuse pigmentation with an overlying ``ground-glass'' white haze or ``veil'' as if the image were out of focus there. 
For discriminability, the pigmentation cannot occupy the entire lesion and is found mainly in the papular part of the lesion.  
On the other hand, regression is defined as areas with white scar-like
depigmentation and/or blue-white to grey pepper-like granules (a.k.a.\
peppering). A particular pitfall when assessing the so-called combinations of
white and blue areas is the fact that this combination is virtually
indistinguishable between the blue-white veil and regression
structures\footnote{Some clinical references use ``blue-white due to
orthokeratosis'' vs.\ ``blue-white due to regression'' to distinguish between
the two features.}. To improve the diagnostic efficacy and increase
inter-observer reproducibility, the two terminologies were unified (although it
was demonstrated that these correspond to two different histopathologic subtrates)
into the definition of blue-whitish structure, during the Consensus Net Meeting on Dermoscopy \cite{argenziano_dermoscopy_2003}. An example of this feature is given in Fig.\ref{FIG:BWS}.

\section{Related Work}
Colour assessment, as discussed, is essential in the clinical diagnosis of skin cancers. Due to this diagnostic importance, many studies have either focused on or employed colour features as a constituent part of their skin lesion analysis systems.  These studies range from employing low-level colour features, such as simple statistical measures of colours occurring in the lesion \cite{madooei_intrinsic_2012,madooei_probabilistic_2014}, to availing themselves of high-level semantic features such as presence of blue-white veil \cite{celebi_automatic_2008,madooei_automatic_2013} or colour variegation \cite{umbaugh_automatic_1989a,seidenari_automated_2006} in the lesion. 

There exists extensive literature on skin lesion image 
analysis (see \cite{korotkov_computerized_2012} for a survey).
However, there are not many studies that report a method specifically pertaining
to the detection of the feature under study here. A handful of studies aim to detect (and localize) blue-white veil in dermoscopy images. Prior work is briefly reviewed next. 

Ogorzalek~et~al.\ \cite{ogorzalek_computational_2010} used thresholding in RGB
space to detect white, black and blue-grey areas. Blue-grey areas were
identified as pixels which satisfied $R>60$, $R-46<G<R+15$ or $G>B-30$,
$R-30<B<R+45$. There is no indication on how these decision rules were
generated. Also, the paper does not provide any experiment to evaluate the
success of colour-based detection. The detected colours were quantified by their
area as part of a feature set for classification of skin lesions (for computer-aided diagnosis).  

Sforza~et~al.\ \cite{sforza_adaptive_2011} proposed an adaptive segmentation of
grey (blue-grey) areas in dermoscopy images. They achieved this by thresholding
on the B component of HSB colour space. The threshold values were induced
``adaptively'', although the paper is unclear on how this adaptive process was
carried out. The paper also lacks quantitative evaluation: results are shown qualitatively for only five dermoscopy images.

Celebi~et~al.\ \cite{celebi_detection_2006,celebi_automatic_2008} detected blue-white veil areas through pixel classification using explicit thresholding, where the threshold values were induced by a trained decision tree. Various colour and texture features were used to characterize the image pixels. The method was tested (for localization accuracy) on 100 dermoscopy images and produced sensitivity of 84\% and a specificity of 96\%. Celebi~et~al.\ further developed a second decision tree classifier to use detected blue-white veil areas for discriminating between melanoma and benign lesions. The detected areas were characterized using simple numerical parameters such as region area, circularity and ellipticity measures. Experiments on a set of 545 dermoscopy images yielded a sensitivity of 69\% and a specificity of 89\%. 

The findings of Celebi et~al.\ indicate
that blue-white veil colour data has a restricted range of values and does not 
fall randomly in a given colour feature space. 
This indicates that their method can benefit from the choice of colour representation.
To investigate this, Madooei~et~al.\ \cite{madooei_automatic_2013} reproduced
Celebi's training experiment where each pixel is represented by its corresponding coordinates 
in various colour spaces. Their investigation revealed that by thresholding over
the Luminance channel ($\text{Lum}=R+G+B$) and normalized blue
($\text{n}B=B/\text{Lum}$), one can obtain equally good results with
considerably less computation compared to \cite{celebi_automatic_2008}. 

Devita~et~al.\ \cite{devita_statistical_2012,fabbrocini_automatic_2014} detected
image regions containing blue-white veil, irregular pigmentation or regression
features. To this aim, first, the lesion was segmented into homogeneous colour regions. Next, simple statistical parameters such as mean and standard deviation were extracted, from HSI colour components, for each region. Finally, a Logistic Model Tree (LMT) was trained to detect each colour feature. LMT is a supervised learning classification model that combines logistic regression and decision tree learning. Devita~et~al.\ detected these colour features as part of their system \cite{fabbrocini_automatic_2014} for automatic diagnosis of melanoma based on the 7-Point checklist clinical guideline. They also evaluated the performance of their colour detection method over a set of 287 images (150 images were used for training and 137 for testing). It is not clear whether the test was aimed to identify (presence/absence) or to localize  the colour features. Nevertheless, results are shown with average specificity and sensitivity of about 80\%.  

Madooei~et~al. \cite{madooei_automatic_2013} identified the blue-white veil
feature in each dermoscopy image through a nearest neighbour matching of image
regions to colour patches of a ``blue-white veil colour palette'': a discrete set of Munsell colours best describing the feature. 
The palette was created by mapping instances of veil and non-veil data to
Munsell colour patches, keeping these colours that exclusively described the feature with highest frequency.
Madooei~et~al.\ claim their method mimics the colour assessment performed by
human observers, pointing to the fact that in identifying a colour, observers are influenced by the colours they saw previously. 
They tested their proposed method for localization of blue-white veil feature on a set of 223 dermoscopy images and reported sensitivity of 71\% and a specificity of 97\%. They also tested their method, in a different experiment, to identify only the presence (or absence) of this feature on a set of 300 images with two subsets of 200 `easy' and 100 `challenging' cases. An image was considered challenging if the blue-white veil area was too small, too pale, occluded, or had variegated colour. They reported accuracy of 87\% and 67\% 
on easy and challenging sets respectively. 

Wadhawan~et~al.\ \cite{wadhawan_detection_2012} detected blue-white veil areas through classification of image patches. The image patches were extracted over the lesion area using a regular-grid sampling. For each image patch, a feature vector was computed by concatenating histogram representation of pixel values in various colour channels of different colour spaces.    
Wadhawan~et~al.\ evaluated their method by performing 10-fold cross-validation
on a set of 489 dermoscopy images (163 containing the veil and remaining 326
free of this feature). For training, images were manually segmented and
annotated by one of the authors. Support vector machine (SVM) with linear kernel
was used for classification. For testing, only presence/absence of the feature
was considered. Results were reported with average sensitivity of about 95\% and average specificity of about 70\%. 

Lingala~et~al.\ \cite{lingala_fuzzy_2014} detected blue areas in dermoscopy
images and further classified them to the three shades lavender, dark and light
blue using fuzzy set membership functions. Their colour detection method builds
on a simple thresholding approach similar to Ogorzalek~et~al.\
\cite{ogorzalek_computational_2010}. A pixel is considered as `blue' if its
normalized RGB values are within a certain range determined empirically (the
threshold values are not reported). These blue areas are further classified into
lavender, light and dark blue by thresholding their intensity value (the luminance channel). 
This thresholding scheme is used to generate training data using 22 dermoscopy images. 
The training data is then used to determine the parameters of fuzzy set
membership functions for three shades of blue. The method is evaluated over a
set of 866 images (173 melanoma and 693 benign). There is no indication of how
successful the colour detection was. Evaluation was conducted by classifying
lesions as melanoma vs.\ benign by extracting simple statistical features over
blue areas. Interestingly, using fuzzy set membership vs.\ simple thresholding
was reported to improve classification performance by less than 0.5\% which
calls into question the effectiveness of the proposed method.

It is to be noted that there are other studies aimed at colour classification
\cite{sboner_knowledge_2001,chen_colour_2003,sboner_multiple_2003,seidenari_computer_2003,pellacani_automated_2004c,sboner_clinical_2004,seidenari_colors_2005,seidenari_early_2006,alcon_automatic_2009,cavalcanti_automated_2011,silva_separability_2012,cavalcanti_two-stage_2013,barata_color_2014}
where the objective is to assign labels (such as colour names) to each region
(or pixel) of the image using the colour information contained in that region.
In some of these, the general colour class of blue or white are considered. This
is, however, different from identifying blue-white structures. We shall remind
the reader that BWS is a specific dermoscopic feature and not necessarily a
particular hue. In fact, what dermatologists annotate as BWS is a mixture of many different hues including various shades of blue, white, grey and sometimes purple.

Also note that the use of the blue-white veil feature has been reported in some 
commercially available computer-aided diagnosis (CAD) systems that do use colour 
information (see e.g.\ \cite{menzies_performance_2005}). These studies however often omit description of methods and techniques which are used for feature extraction (perhaps due to patent protection) and therefore were not considered here.

\textbf{Summary -- }
Most of the prior works are pixel-based classification approaches where the
classification method often is simply to partition a colour space by imposing a
set of decision boundaries (thresholds) either found empirically or induced by
e.g.\ training a decision tree (see Table.\ref{tab:colour_detection}).  
Since dermoscopic features are in fact defined over a region of at least 10\% of
the size of the lesion, pixel-based classification would seem to be an inappropriate approach. Also, caution must be taken in defining the decision boundaries (threshold values): the colour values are highly dependent on the acquisition technique and 
the imaging setup. For example, if illumination changes, or in cases with shadows, shading, and 
other colour-degrading factors, thresholding methods might fail ungracefully.

\begin{table}[h]
\centering
\caption{Summary of related works.}
\begin{tabular}{llll}
\toprule
Author                                                & Year & Method              & Approach     \\
\midrule
Celebi~et~al.\ \cite{celebi_detection_2006}           & 2006 & Decision tree       & Pixel-based  \\
Celebi~et~al.\ \cite{celebi_automatic_2008}           & 2008 & Decision tree       & Pixel-based  \\
Ogorzalek~et~al.\ \cite{ogorzalek_computational_2010} & 2010 & Thresholding        & Pixel-based  \\
Sforza~et~al.\ \cite{sforza_adaptive_2011}            & 2011 & Thresholding        & Pixel-based  \\
Devita~et~al.\ \cite{devita_statistical_2012}         & 2012 & LMT                 & Region-based \\
Wadhawan~et~al.\ \cite{wadhawan_detection_2012}       & 2012 & SVM                 & Region-based \\
Madooei~et~al\ \cite{madooei_automatic_2013}          & 2013 & Thresholding        & Pixel-based  \\
Madooei~et~al\ \cite{madooei_colour_2013}             & 2013 & Colour Palette      & Region-based \\
Fabbrocini~et~al.\ \cite{fabbrocini_automatic_2014}   & 2014 & LMT                 & Region-based \\
Lingala~et~al.\ \cite{lingala_fuzzy_2014}             & 2014 & Fuzzy sets          & Pixel-based  \\ 
\bottomrule
\end{tabular}
\label{tab:colour_detection}
\end{table}

In all the studies reported here, the emphasis is to use colour features, and
structural information such as texture are either ignored
\cite{ogorzalek_computational_2010,wadhawan_detection_2012,devita_statistical_2012,fabbrocini_automatic_2014,lingala_fuzzy_2014}
or found to be not useful \cite{celebi_automatic_2008,madooei_automatic_2013}. This is problematic since these detectors would potentially fail to distinguish between a BWS and a similar feature in a benign lesion.\footnote{Uniform blue-white structures may be observed in common moles such as blue nevi, but in melanoma they are diffuse, asymmetric and irregular.} 
Moreover, these studies fall into the classical paradigm of supervised learning
that requires fully annotated data. However, this exhaustive labeling approach
is costly and error prone, especially since such annotations have been always
made by a single expert rather than via a consensus of experts' opinions.
It is hard to make outright claims about the success of these algorithms,
especially since often these studies have failed to provide comparisons to other algorithms.

Another caveat appears here:
the BWS, if identified correctly, is highly specific to melanoma. 
However, it is not its presence that is the diagnostic indicator, but rather the
extent to which it manifests in a lesion with respect to other dermoscopic characteristics of the lesion.
The blue and whitish area are also found in benign lesions. However, in combination with other dermoscopic features such as atypical network or irregular globules, these colours are diagnosed as a malignancy criteria.
This imposes an extra challenge since any computer program using localization of this feature should be (somehow) aware of other local features.

One possible solution is to use a Structured Prediction paradigm, which allows 
training of a classifier for structured output labels. The output can be a set of dermoscopic criteria (including the BWS and other associated features) and e.g.\ a graphical model can be used to learn the relationship (structure) between the labels. 
We hypothesize, in the case of our alternative MIL-based approach, that since the BWS instances are not annotated in our image set, the detector would learn to recognize those \textit{salient} regions that contain BWS in association with e.g.\ pigment network alterations, irregular globules, etc. Results from our experiments (\S\ref{experiment}) support this hypothesis. 

\section{Proposed Method}\label{mil}

\begin{figure*}[ht]
\begin{center}
\includegraphics[width=0.8\linewidth]{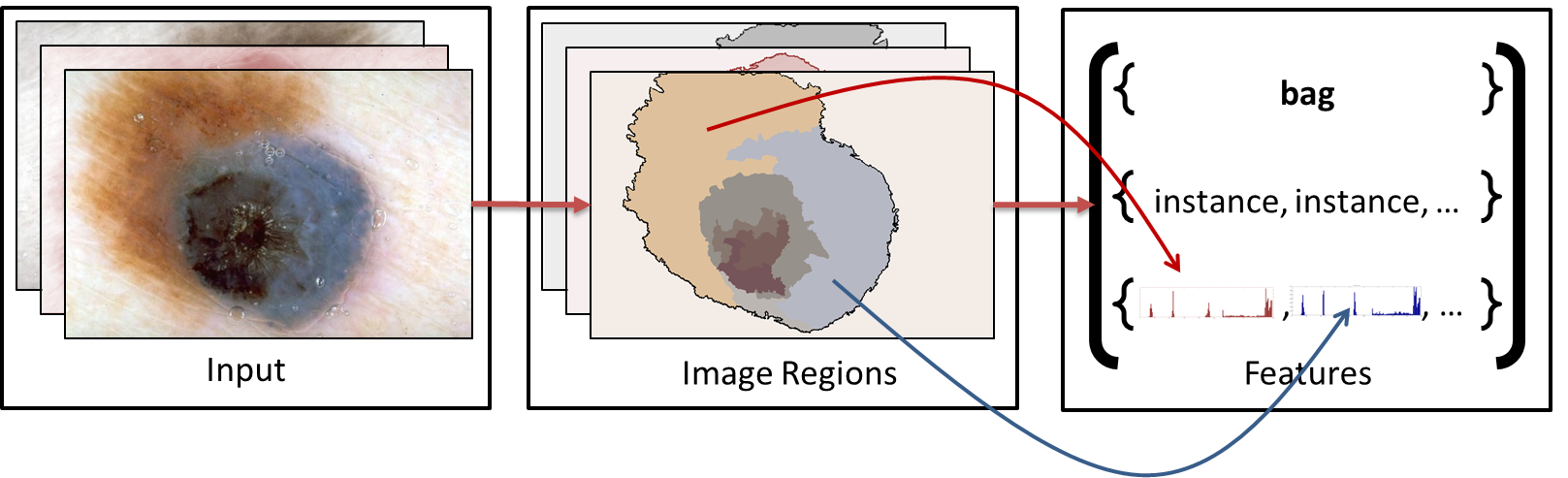}
\end{center}
\caption{Schematic representation of data representation in our proposed model.}
\label{fig:data}
\end{figure*}

Our goal is to learn a detector for BWS from a set of images, each with a binary
label $Y \in \lbrace +1, -1 \rbrace$ indicating presence or absence of the BWS feature.  
We model an image as a set of non-overlapping regions $\mathbf{X}=\lbrace \mathbf{x}_{1}, \mathbf{x}_{2}, \cdots, \mathbf{x}_{m} \rbrace$
where regions also have unknown binary labels $y_{i} \in \lbrace +1, -1 \rbrace$ (the region may or may not correspond to an instance of BWS). This reduces the problem of BWS localization to the problem of binary classification of image regions.

We used the well-known mean-shift segmentation tool EDISON \cite{georgescu_edge_2003} to extract image regions.\footnote{EDISON parameters: SpatialBandWidth=7, RangeBandWidth=6.5,
and MinimumRegionArea=0.01*nRows*nColumns.} 
To further reduce the number of instances, we discard regions outside the lesion area. 
The lesion is detected by application of grey-level thresholding method of Otsu
\cite{otsu_threshold_1979} (classical foreground/background detection) as
described in \cite{madooei_automated_2012}. Note that neither region nor lesion
segmentation are main constituents of our approach; we could instead use e.g.\ a regular grid of windows over the image.

Each region is represented by a fixed-size feature vector $\mathbf{x}_{i} = \left[ x_{i1}, \cdots, x_{iD} \right] \in \mathbb{R}^D$. 
Our choice of feature was a concatenated (and normalized) histogram of colour and texture information extracted over each region. For texture, we used two popular descriptors from the texture classification literature: LBP \cite{ojala_multiresolution_2002} and MR8 \cite{varma_statistical_2005}.  

Colour features are the most important information to be captured here. We used a uni-dimensional histogram of CIE~Lab colour values.  
CIE~Lab is a perceptual colour space based on opponent process theory. A
property of this space is that distances correspond (approximately) linearly to
human perception of colour. This is desired since we apply uniform binning (in
construction of colour histograms), which implicitly assumes a meaningful distance measure. 
The colour histogram is constructed using a bin size of 5 units. Thus each bin has a radius of $\sim 1$ JND\footnote{Weber's Law of Just Noticeable Difference (JND), see \url{http://apps.usd.edu/coglab/WebersLaw.html}} and so it subdivides colour space near the theoretical resolution of human colour differentiation. 

Finally the whole set of training data is represented by $\lbrace \left(
\mathbf{X}_1,Y_1 \right), \cdots, \left( \mathbf{X}_N,Y_N \right) \rbrace$ (a
schematic representation is given in Fig.\ref{fig:data}). At training time we
are only given image-level labels $Y$, leading to the classic MIL problem.

In the MIL setting, training examples are presented as labeled bags (i.e.\ sets) of instances, and the instance labels are not given. According to the standard MIL assumption, a bag is positive if at least one of the instances is positive, while in a negative bag all the instances are negative. 
This assumption fits well with our problem. We can think of each image as a ``bag'' of instances (image regions) where the binary image label $Y=+1$ specifies that the bag contains at least one instance of the BWS feature. The label $Y=-1$ specifies that the image contains no instances of the feature. 

MIL problems are typically solved (locally) by finding a local minimum of a non-convex objective function, such as mi-SVM \cite{andrews_support_2002}. 
In this paper, we use a recent MIL algorithm, the multi-instance Markov network (MIMN)~\cite{hajimirsadeghi_multiple_2013}. 
It is proved in \cite{hajimirsadeghi_multiple_2013} that MIMN is a generalized version of
mi-SVM with guaranteed convergence to an optimum (unlike mi-SVM, which might get stuck in a loop and never converge). 

This method introduces a probabilistic graphical model for multi-instance classification. Because of the multi-unit and structural nature of probabilistic graphical models, they seem to be powerful tools for MIL. 
The proposed algorithm works by parameterizing a cardinality potential on latent instance labels in a Markov network. Consequently the model can deal with different levels of ambiguity in the instance labels and model the standard MIL assumption as well as more generalized MIL assumptions. 
On the other hand, this graphical model leads to principled and efficient inference algorithms for predicting both the bag label and instance labels.

\begin{figure}[b]
\begin{center}
\includegraphics[width=0.6\linewidth]{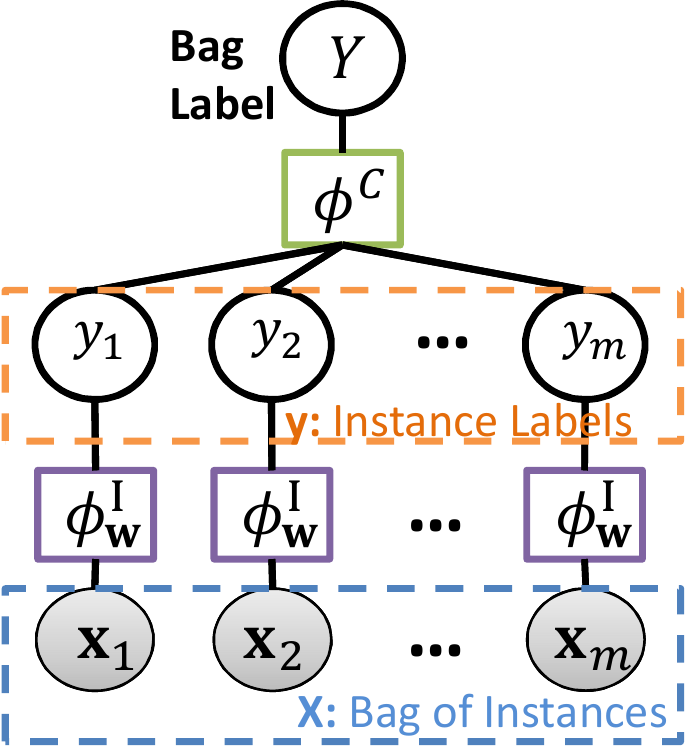}
\end{center}
\caption{Graphical illustration of the MIMN model for MIL. Instance potential functions $\phi^I_{\mathbf{w}}$ relate instances $\mathbf{x}_i$ to labels $y_i$.  A clique potential $\phi^C$ relates all instance labels $y_i$ to the bag label $Y$.}
\label{fig:model}
\end{figure}

The graphical representation of the MIMN model is shown in Figure~\ref{fig:model}. Given this model, a scoring function over tuples $(\mathbf{X},\mathbf{y},Y)$ is defined as:
\begin{equation}\label{eq:modelPot}
f_{\mathbf{w}}(\mathbf{X}, \mathbf{y}, Y) =  \phi^C(\mathbf{y}, Y) + \sum_i{\phi^I_{\mathbf{w}}(\mathbf{x}_i, y_i)}
\end{equation}

This Markov network consists of the instance-label potentials $\phi^I_{\mathbf{w}}(\mathbf{x}_i, y_i)$ and a cardinality clique potential $\phi^C(\mathbf{y}, Y)$. The instance-label potentials are parameterized as:
\begin{equation}\label{eq:ilPot}
\phi^I_{\mathbf{w}}(\mathbf{x}_i, y_i) = \mathbf{w}_{I}^{\top} \mathbf{x}_i \, y_i
\end{equation}
and the cardinality clique potential is parameterized by two different cardinality functions, one for positive bags ($C^{+}$) and one for negative bags ($C^{-}$):
\begin{equation}\label{eq:labelsPot}
\begin{split}
\phi^C(\mathbf{y}, Y) & = C\left(m^{+}, m^{-}, Y \right) \\
& = C^{+}\left(m^{+}, m^{-} \right) \mathds{1}(Y = 1) \\ 
& + C^{-}\left(m^{+}, m^{-} \right) \mathds{1}(Y = -1),
\end{split}
\end{equation}
where $m^{+}$/$m^{-}$ denotes the number of instance labels in $\mathbf{y}$ which are inferred to be positive/negative. By appropriate parameterization of $C^{+}$ and $C^{-}$, the standard MIL assumption can be modeled:
\begin{flalign}
& C^{+}(0, m) = -\infty \label{eq:mimn1} \\
& C^{+}(m^+, m - m^+) = 0 \quad m^+ = 1, \cdots, m \label{eq:mimn2}\\
& C^{-}(0, m) = 0 \label{eq:mimn3} \\
& C^{-}(m^+, m - m^+) = -\infty \quad m^+ = 1, \cdots, m \label{eq:mimn4}
\end{flalign}
This formulation encodes that there must be at least one positive instance in a
positive bag (\ref{eq:mimn1})\&(\ref{eq:mimn2}). However, there must not be any positive instances in a negative bag (\ref{eq:mimn3})\&(\ref{eq:mimn4}).

\textbf{Inference --}
Given the MIMN model, the inference problem is to find the instance labels, by solving the following optimization problem: 
\begin{equation}\label{eq:h-infer}
F_{\mathbf{w}}(\mathbf{X}, Y) = \max_{\mathbf{y}} f_{\mathbf{w}}(\mathbf{X}, \mathbf{y}, Y).
\end{equation}
It was shown in~\cite{hajimirsadeghi_multiple_2013} how to efficiently solve this inference problem in $O(m \log m)$ time. Next, the bag label can be predicted by simply running inference twice, trying $Y = +1$ and $Y = -1$ and taking the label which maximizes $F_{\mathbf{w}}$.

\textbf{Learning -- }
Similar to the relations in latent SVM, the learning problem is formulated in a max-margin discriminative framework by minimizing a regularized hinge loss function:
\begin{equation}\label{eq:learning}
\begin{split}
& \min_{\mathbf{w}}{\sum\limits_{n = 1}^{N}{\left( \mathcal{L}_n - \mathcal{R}_n \right)} + \frac{\lambda}{2} \lVert \mathbf{w} \rVert^2} \\
& \text{where } \mathcal{L}_n = \max_Y{\max_{\mathbf{y}}{\left(\Delta(Y, Y_n) + f_{\mathbf{w}}(\mathbf{X}_n, \mathbf{y}, Y)\right)}}, \quad \\
& \mathcal{R}^n = \max_{\mathbf{y}}{f_{\mathbf{w}}(\mathbf{X}_n, \mathbf{y}, Y_n)}, \\
& \Delta(y, y_n) = \begin{cases}
1 & \text{if } Y \neq Y_n \\
0 & \text{if } Y = Y_n.
\end{cases}
\end{split}
\end{equation}
This can be solved by using the non-convex cutting plane method in \cite{do_large_2009}.

\section{Experiments and Results}\label{experiment}
We make use of a set of dermoscopy images from the CD-ROM Interactive Atlas of
Dermoscopy \cite{argenziano_interactive_2000} (from now on denoted Atlas for brevity). This educational media contains a
collection of about 1000 clinical cases acquired in three institutions in
Europe. All cases are accompanied with clinical data including dermoscopic
images, diagnosis (nearly all confirmed histo-pathologically), and consensus
documentation of dermoscopic criteria. Thus, all images are weakly-labeled. That
is, although there are image-level labels that encode whether an image contains
a dermoscopic feature or not, the features are not locally annotated. The
dataset is a well-known (de~facto) benchmark, albeit most studies use only a small subset since for typical supervised learning methods manual labelling is required.

The proposed method is tested on a set of 855 dermoscopy images selected from
the Atlas. Images were excluded if the lesion was heavily
occluded by hair or oil artifacts, or if they were located on palms/soles, lips
or genitalia. In our selected set, 155 images are documented to contain
blue-white veil regions, 156 images contain blue (or combination of white and
blue) regression structures, and 43 images contain both of these dermoscopic
criteria (thus a total of 354 positive BWS cases). The remaining 501 are free of these features.  
We consider this set as challenging since not many images contain a sizeable BWS and most others on the other hand contain too small, too pale, occluded, or variegated colour BWS instances. Also note that there are various other dermoscopic features present in each image. 

Table.\ref{tab:results} reports the results over 3-fold cross validation for the
main task of BWS identification (i.e.\ whether the image contains the feature or
not). For comparison, we considered the most prominent studies amongst the prior
arts: Celebi~et~al.\ \cite{celebi_automatic_2008} and Madooei~et~al.\
\cite{madooei_automatic_2013}. These studies report a method, experimental
procedure, and results specifically pertaining to the detection of the feature under study here.
Compared to prior work, our method shows substantial improvements with specificity boosted by $\sim 26\%$, precision increased by $\sim 13\%$, and accuracy improved by $\sim 7\%$. The f-score of our detector is comparable to that of the prior art. Our method's recall lags behind that of \cite{celebi_automatic_2008,madooei_automatic_2013}. We would like to bring
the readers attention to the gap between the precision and recall for baseline
methods of \cite{celebi_automatic_2008,madooei_automatic_2013}; their good
recall is achieved at the expense of high false positives. Our method on the other hand maintains a steady performance level. 

Note that both \cite{celebi_automatic_2008,madooei_automatic_2013} use only
colour features. They found that texture information was not useful; but here
indeed our MIL-based method improves by adding texture. We used the same texture
features \cite{celebi_automatic_2008,madooei_automatic_2013}. This makes sense,
and further demonstrates the capacity of MIL to make use of such information
towards the computational task at hand. Even using only colour, our proposed method still outperforms \cite{celebi_automatic_2008,madooei_automatic_2013} by a large margin. For comparison, we have added a row to Table.\ref{tab:results} with performance measures using only colour features.    

\begin{table*}[htbp]
  \normalsize
  \centering
  \caption{BWS detection : Proposed method vs.\ \cite{celebi_automatic_2008,madooei_automatic_2013} }
    \begin{tabular}{llccccc}
    \toprule
    Dataset & Method & Accuracy & Precision & Recall & f-score & Specificity\\
    \midrule
    \multirow{4}{*}{Atlas \cite{argenziano_interactive_2000}} & Proposed method$^{a}$ & \textbf{72.63} & \textbf{68.07}  & 63.84          & 65.89           & \textbf{78.84} \\
                          & Proposed method (colour)$^{a,b}$                      & 70.52          & 64.32           & 64.68          & 64.89           & 76.25 \\
                          & Celebi~et~al.\ \cite{celebi_automatic_2008}         & 59.88          & 50.89           & \textbf{88.42} & 64.60           & 39.72 \\
                          & Madooei~et~al.\ \cite{madooei_automatic_2013}       & 65.96          & 55.87           & 84.75          & \textbf{67.34}  & 52.69 \\ \midrule
    \multirow{3}{*}{PH2 \cite{mendonca_ph2_2013}}   & Proposed method$^{c}$           & \textbf{84.50} & \textbf{61.54}  & 74.42          & \textbf{67.37}  & \textbf{87.90} \\
                          & Celebi~et~al.\ \cite{celebi_automatic_2008}$^{c}$         & 79.50          & 51.28           & 93.02          & 66.12           & 75.80 \\
                          & Madooei~et~al.\ \cite{madooei_automatic_2013}$^{c}$       & 76.50          & 47.67           & \textbf{95.35} & 63.57           & 71.43 \\ 
    \bottomrule
    \end{tabular}%
  \label{tab:results}%
  \caption*{$^{a}$Results over 3-fold cross validation. $^{b}$The proposed method trained using only colour features. $^{c}$The method is trained using the Atlas and tested on the PH2 set. Please see \S\ref{experiment} for details and discussion.}
\end{table*}%

Moreover, both \cite{celebi_automatic_2008,madooei_automatic_2013} are
supervised methods and require annotated training data (images with instances of
BWS localized on them) whereas our data is only weakly labelled. Note that we used the detection methods originally produced by \cite{celebi_automatic_2008,madooei_automatic_2013} and did not train these systems again. 
Please refer to Alg.\ref{ALG:BWV-Celebi} and Alg.\ref{ALG:BWV-ColorPalette} for a summary.
We used the code and data of 
\cite{madooei_automatic_2013}, and the implementation in
\cite{madooei_automatic_2013} of \cite{celebi_automatic_2008}. Note that both
\cite{celebi_automatic_2008,madooei_automatic_2013} were originally trained on a
subset of the same dataset that is used here. In our experiment, to compare to
\cite{celebi_automatic_2008,madooei_automatic_2013} we simply run
that code on the whole dataset. One might argue it is unfair to the present
paper since the test data contains their training data as well, whereas our
result is obtained over cross validation with separated test and training sets. 
For further clarification, a short description of the (original) training process of the baseline methods follows. 

\begin{algorithm}
\caption{ -- The method of Celebi~et~al. \cite{celebi_automatic_2008}}
\label{ALG:BWV-Celebi}
\begin{algorithmic}[1]
\STATE {Load a dermoscopy image of skin lesion.}
\STATE {Extract lesion border.}
\STATE {Dilate the border by 10\% of its area.}
\STATE {Extract region outside the dilated border of size 20\% of lesion area.}
\FOR {each pixel in extracted region}
\IF {$R>90$ \AND $R>B$ \AND $R>G$}
\STATE {Mark the pixel as \textit{healthy} skin.}
\ELSE 
\STATE {Ignore the pixel and continue.}
\ENDIF
\ENDFOR 
\STATE {Set $\bar{R_s}$ as the mean of red channel values for pixels marked \textit{healthy} skin.}
\FOR {each pixel in the image}
\STATE {$nB=B/R+G+B$}
\STATE {$rR=R/\bar{R_s}$}
\IF {$nB\geq 0.3$ \AND $-194 \leq rR < -51$}
\STATE {Classify pixel as BWS}
\ENDIF
\ENDFOR 
\end{algorithmic}
\end{algorithm}

\begin{algorithm}
\caption{ -- The method of Madooei~et~al.\ \cite{madooei_automatic_2013}}
\label{ALG:BWV-ColorPalette}
\begin{algorithmic}[1]
\STATE{\underline{\textbf{PART1: Colour Palette}}}
\FOR {each image in database}
\STATE {Convert from sRGB to CIELAB}
\STATE {Replace each pixel with superpixel representation}
\FOR {each pixel marked as veil}
\STATE {Compute the approximate Munsell specification}
\ENDFOR
\ENDFOR
\STATE {Create frequency table from the computed Munsell colour patches, keep the most representative colours (in terms of highest frequency)
and organize them in a palette.}
\STATE \underline{{\textbf{PART2: Detection}}}
\STATE {Load a skin lesion image }
\STATE {Convert from sRGB to CIELAB}
\STATE {Segment using EDISON \cite{georgescu_edge_2003}}
\FOR {each segmented region}
\STATE {Find the best match from colour palette}
\IF {The best match is within the threshold distance}
\STATE {Classify as BWS}
\ENDIF
\ENDFOR 
\end{algorithmic}
\end{algorithm}

\textbf{Celebi~et~al.\ \cite{celebi_automatic_2008}} used a set of 105 dermoscopy images (selected from the Atlas) consisting of 43 images containing sizeable blue-white veil areas with the remaining 62 free of this feature. For each image, a number of small circular regions that either contained the feature or was free of it were manually determined by a dermatologist and used for training. A decision tree classifier with C4.5 \cite{quinlan_c4.5:_1993} induction algorithm was employed to classify each pixel in the training stage into two classes: blue-white veil and otherwise. Among the 18 different colour and texture features included,\footnote{The description of features -- as well as the feature extraction
process-- is omitted for space considerations. The interested reader is referred to \cite{celebi_automatic_2008} for details.} only two features appeared in the induced decision rules: The classification was conducted by thresholding on a normalized-blue channel ($B/\{R+G+B\}$) and relative-red feature (defined as $R-\bar{R}_s$ where $\bar{R}_s$ is the mean of red channel values for \textit{healthy} skin areas only).

\textbf{Madooei~et~al.\ \cite{madooei_automatic_2013}} used the same 105 dermoscopy images employed by \cite{celebi_automatic_2008}. They mapped each colour of blue-white veil data to its closest colour patch in the Munsell system (using the nearest neighbour searching technique).
Interestingly, the 146,353 pixels under analysis mapped to only 116 of the totality of
2352 Munsell colour patches available in their look-up table.\footnote{For implementation details on e.g.\ colour transformation or segmentation parameters, please refer to \cite{madooei_automatic_2013}.} Among these, 98\%
of the veil data was described by only 80 colour patches. These 80 colours were
organized on a palette as a discrete set of Munsell colours best describing the
feature. Madooei~et~al.\ also analyzed non-veil data by the same principle. The
254,739 pixels from non-veil areas mapped to 129 Munsell colour patches, among
which only 3 patches were overlapping with the 116 veil patches. These 3
contribute (all together) to less than 2\% of veil data and were not considered
among the 80 patches in the blue-white veil colour palette. For testing, the blue-white veil feature was segmented in each dermoscopy image through a nearest neighbour matching of image regions to colour patches of their ``blue-white veil palette''. 

\begin{figure*}[ht]
\centering
\subfigure[Input image]{\includegraphics[width=0.22\textwidth]{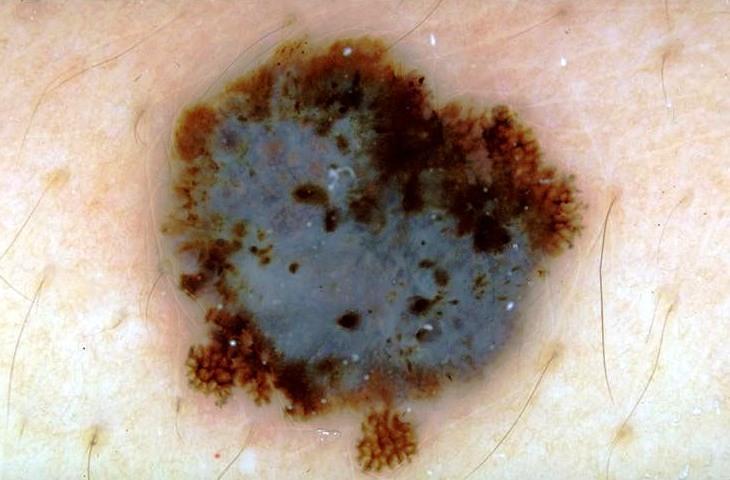} }
\subfigure[Proposed method]{\includegraphics[width=0.22\textwidth]{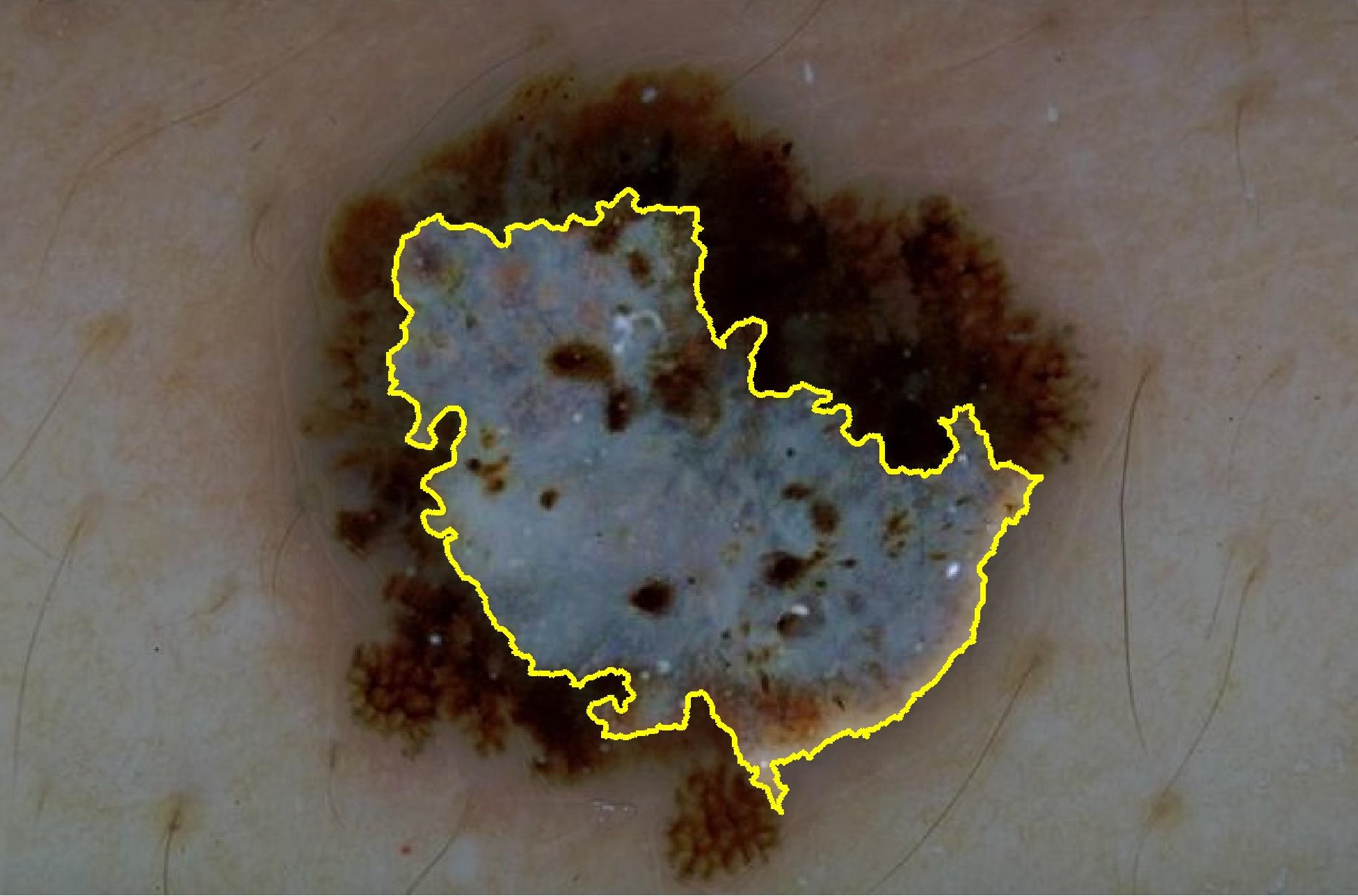} }
\subfigure[Celebi~et~al.\ \cite{celebi_automatic_2008}]{\includegraphics[width=0.22\textwidth]{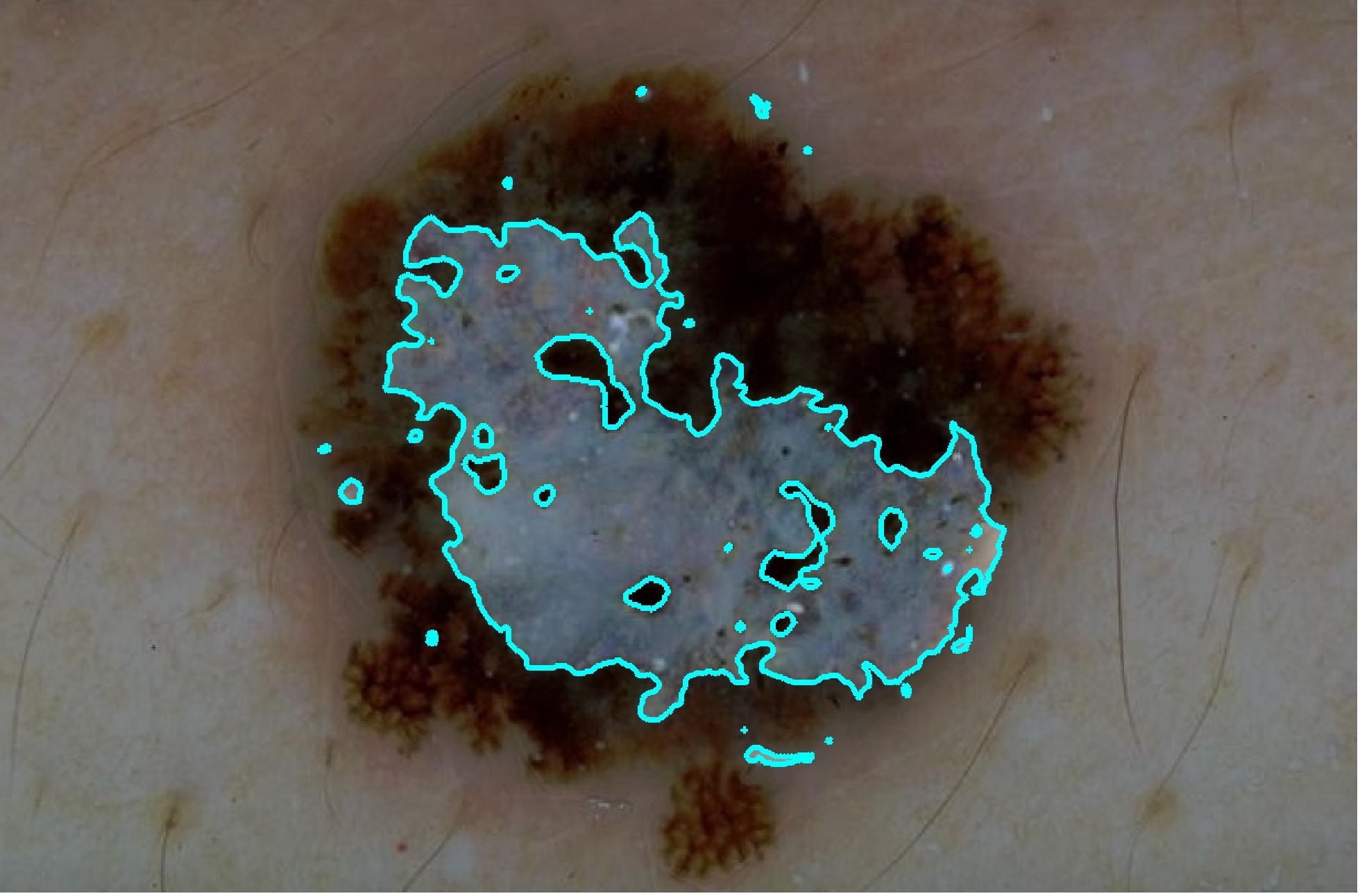} }
\subfigure[Madooei~et~al.\ \cite{madooei_automatic_2013}]{\includegraphics[width=0.22\textwidth]{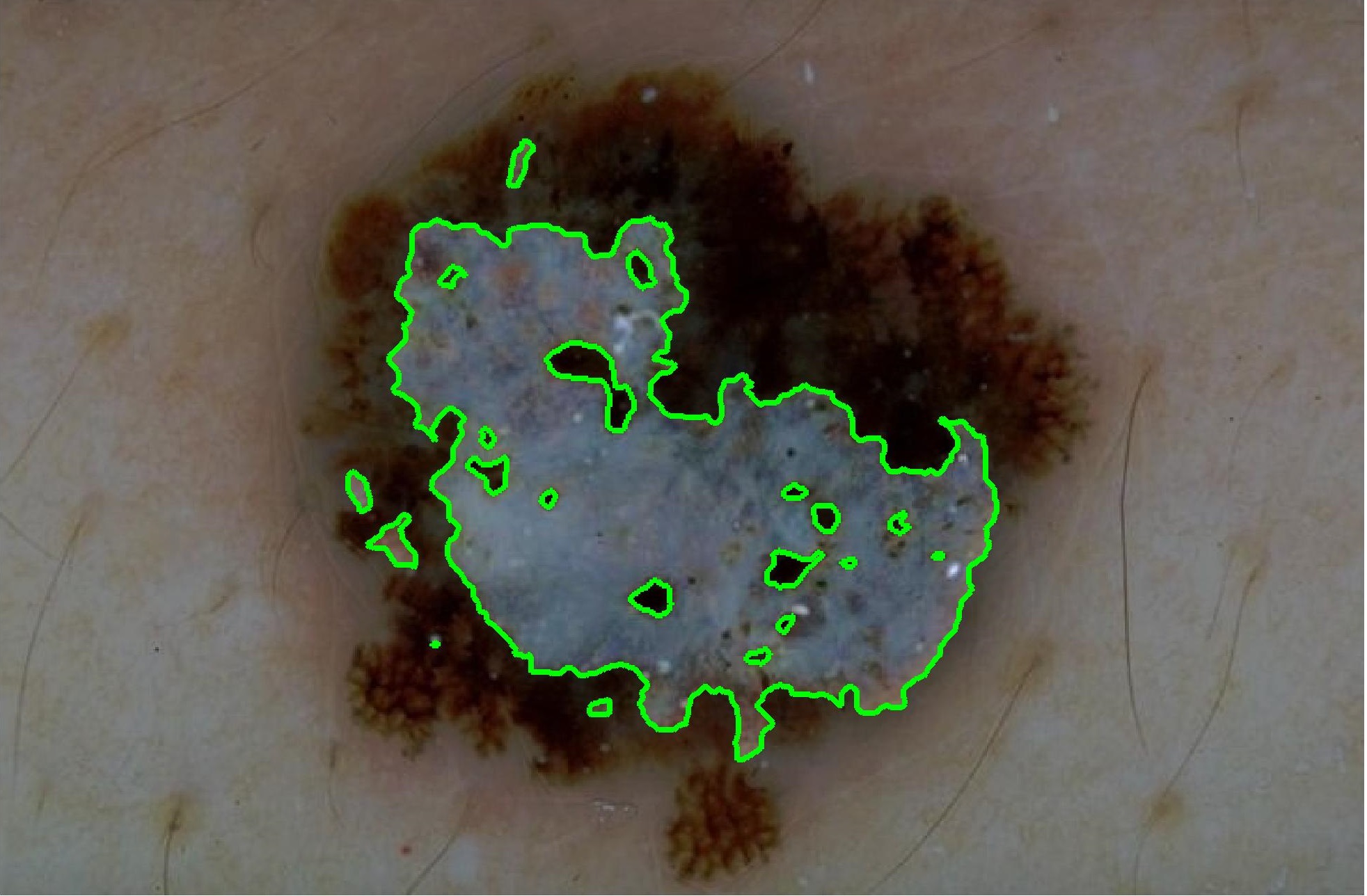} }
\\
\subfigure[]{\includegraphics[width=0.22\textwidth]{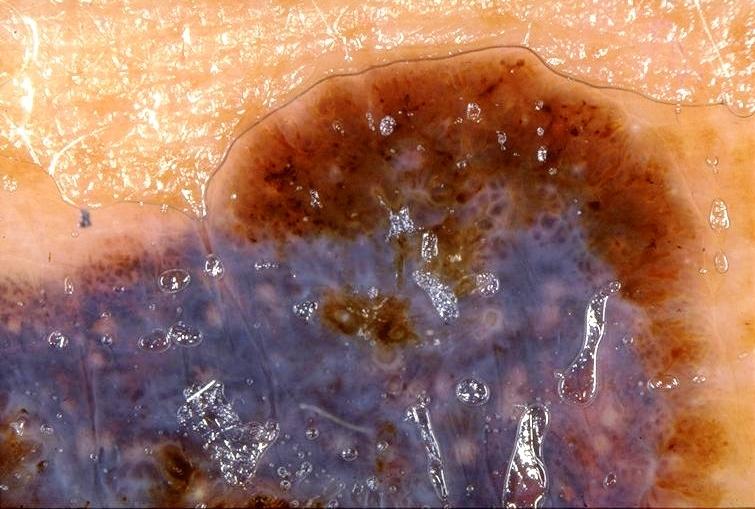} }
\subfigure[]{\includegraphics[width=0.22\textwidth]{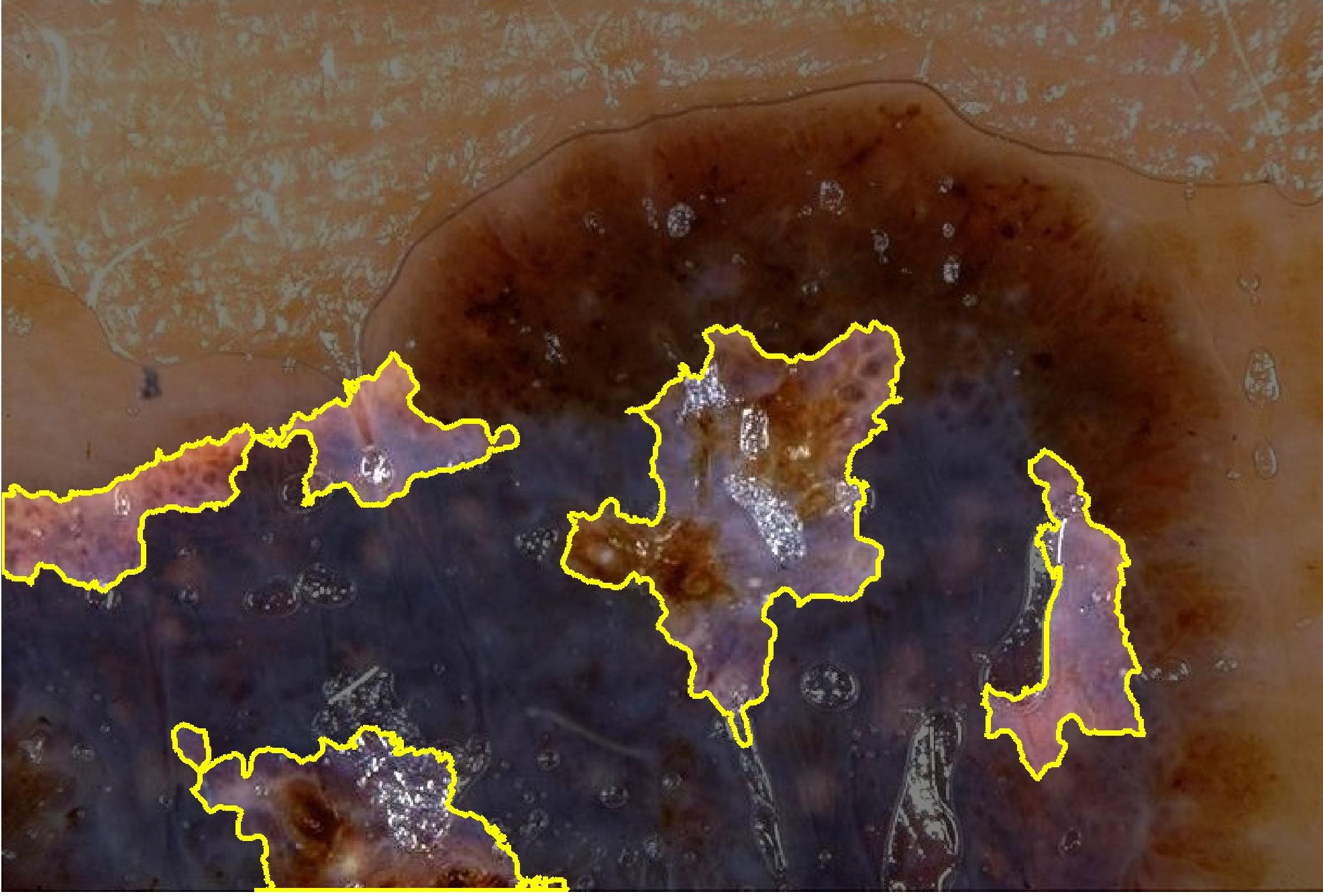} }
\subfigure[]{\includegraphics[width=0.22\textwidth]{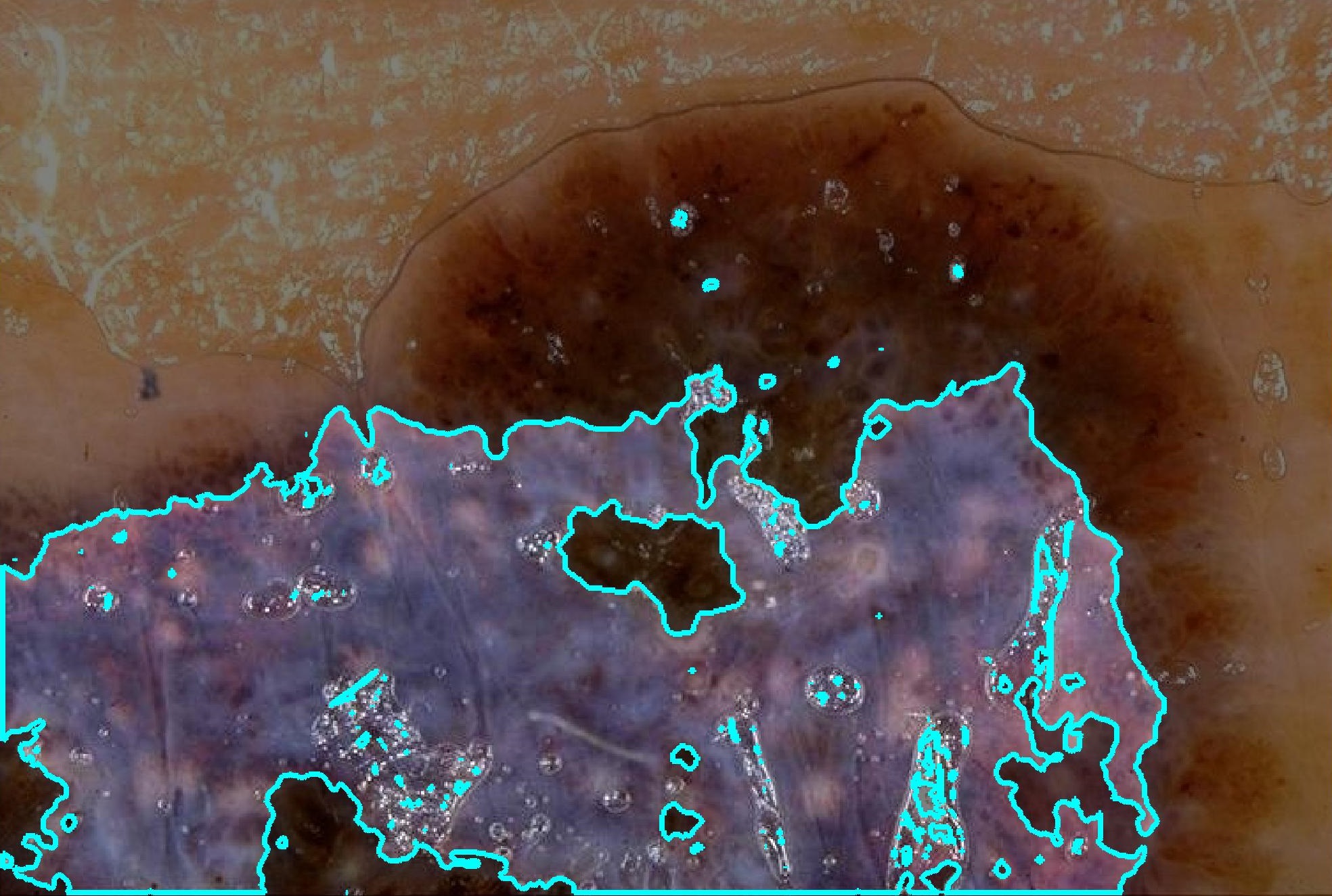} }
\subfigure[]{\includegraphics[width=0.22\textwidth]{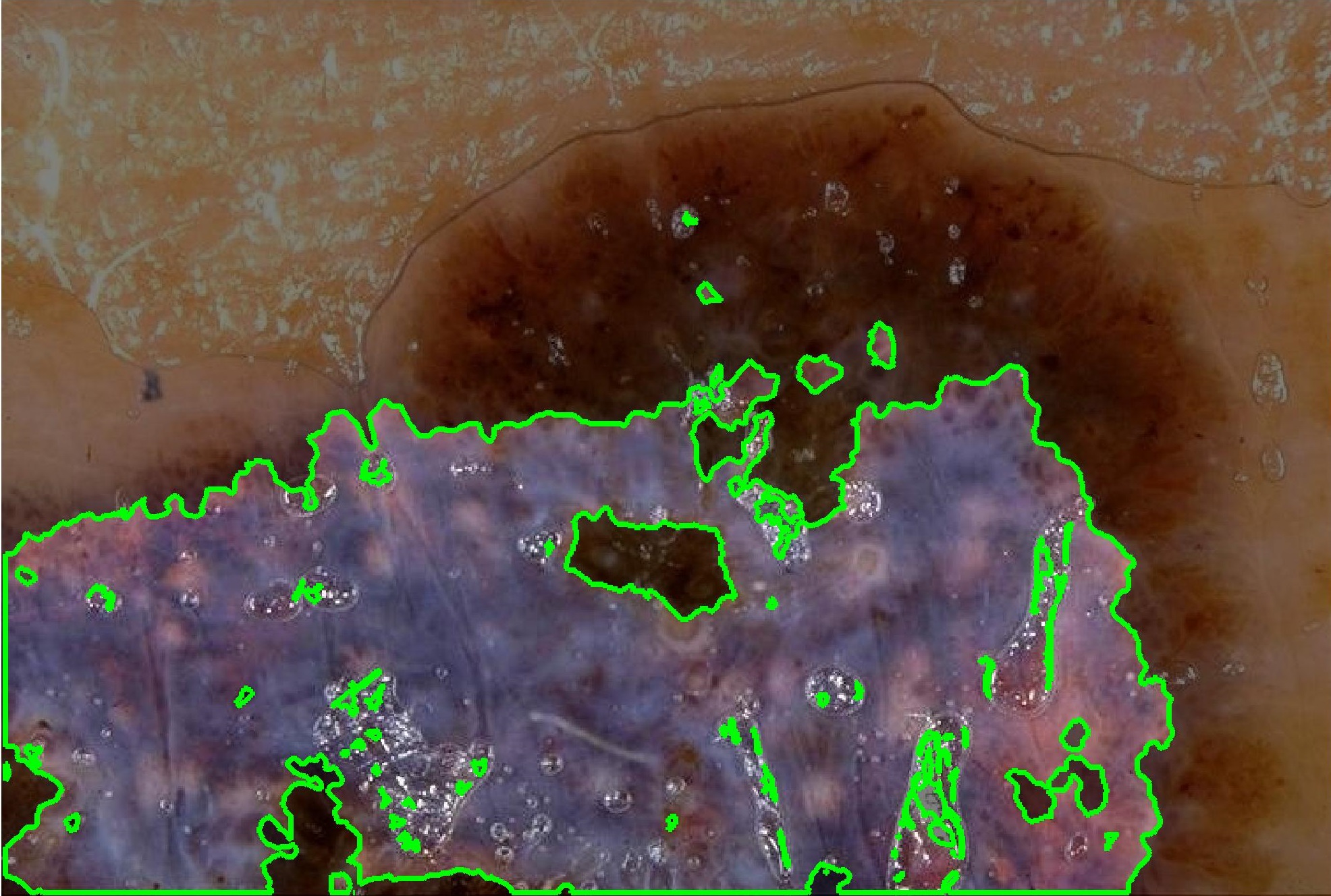} }
\\
\subfigure[]{\includegraphics[width=0.22\textwidth]{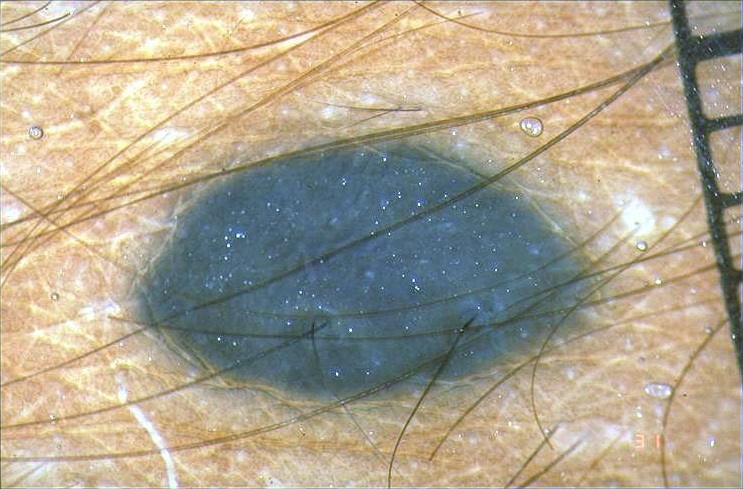} }
\subfigure[]{\includegraphics[width=0.22\textwidth]{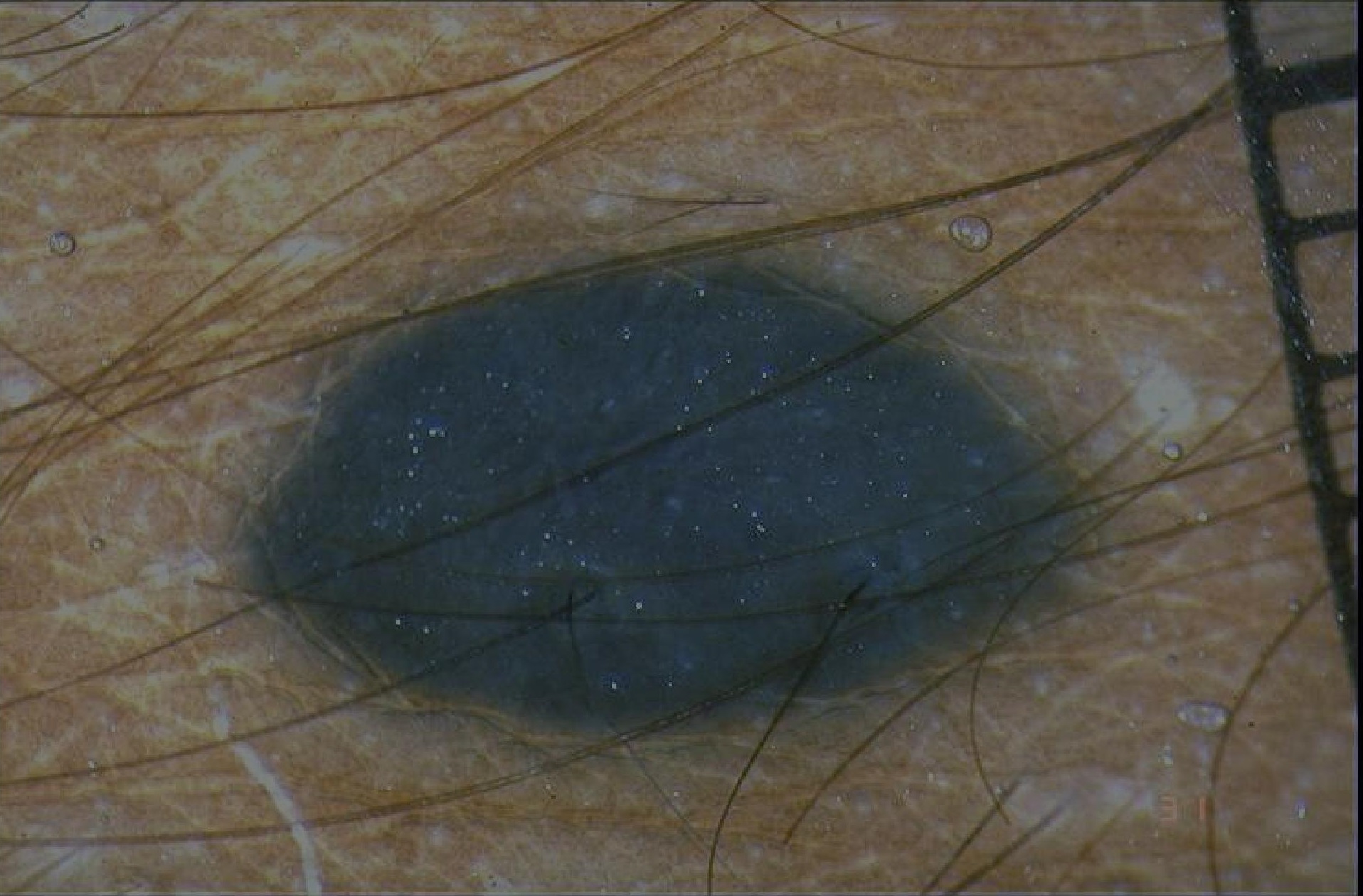} }
\subfigure[]{\includegraphics[width=0.22\textwidth]{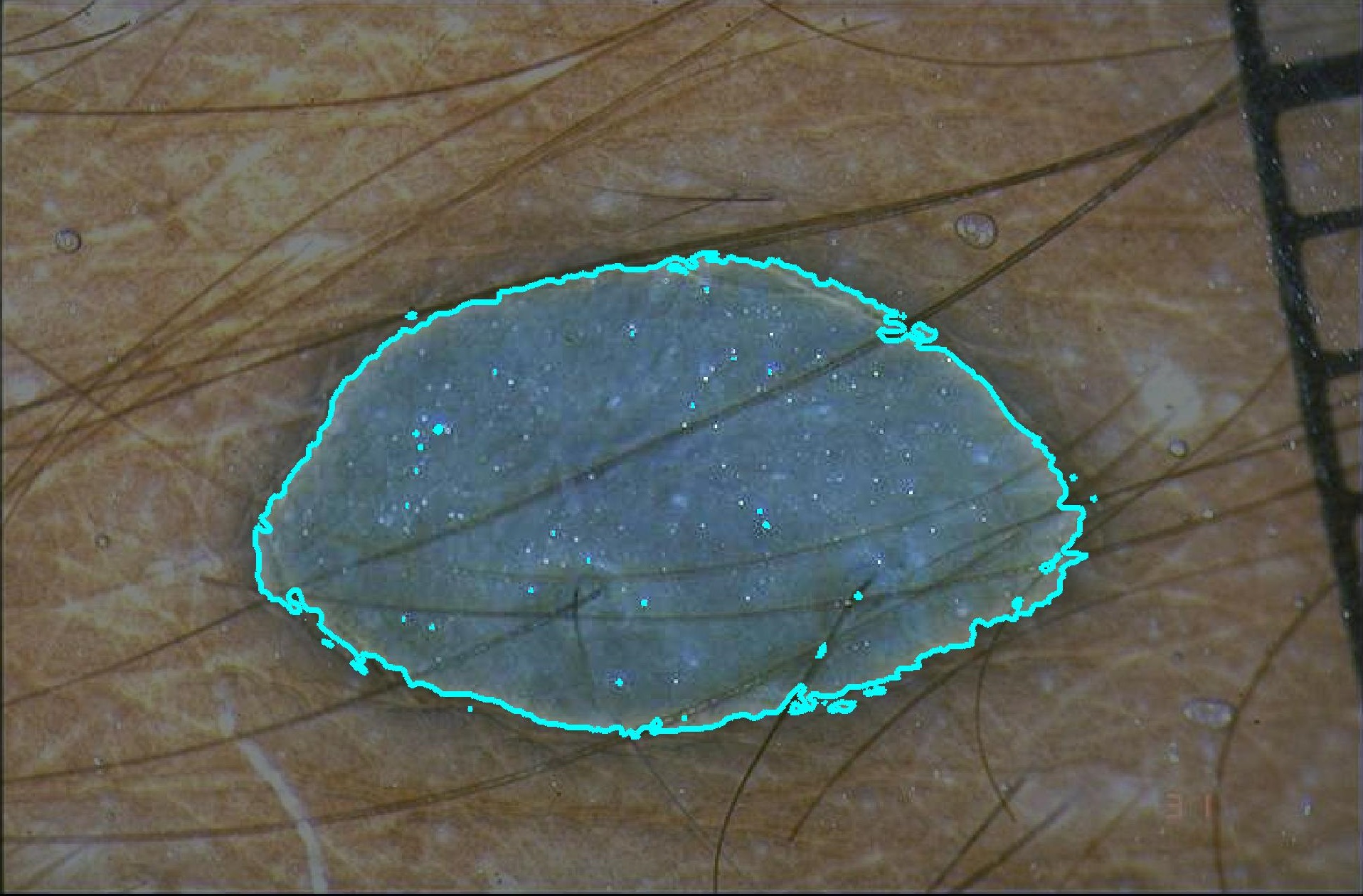} }
\subfigure[]{\includegraphics[width=0.22\textwidth]{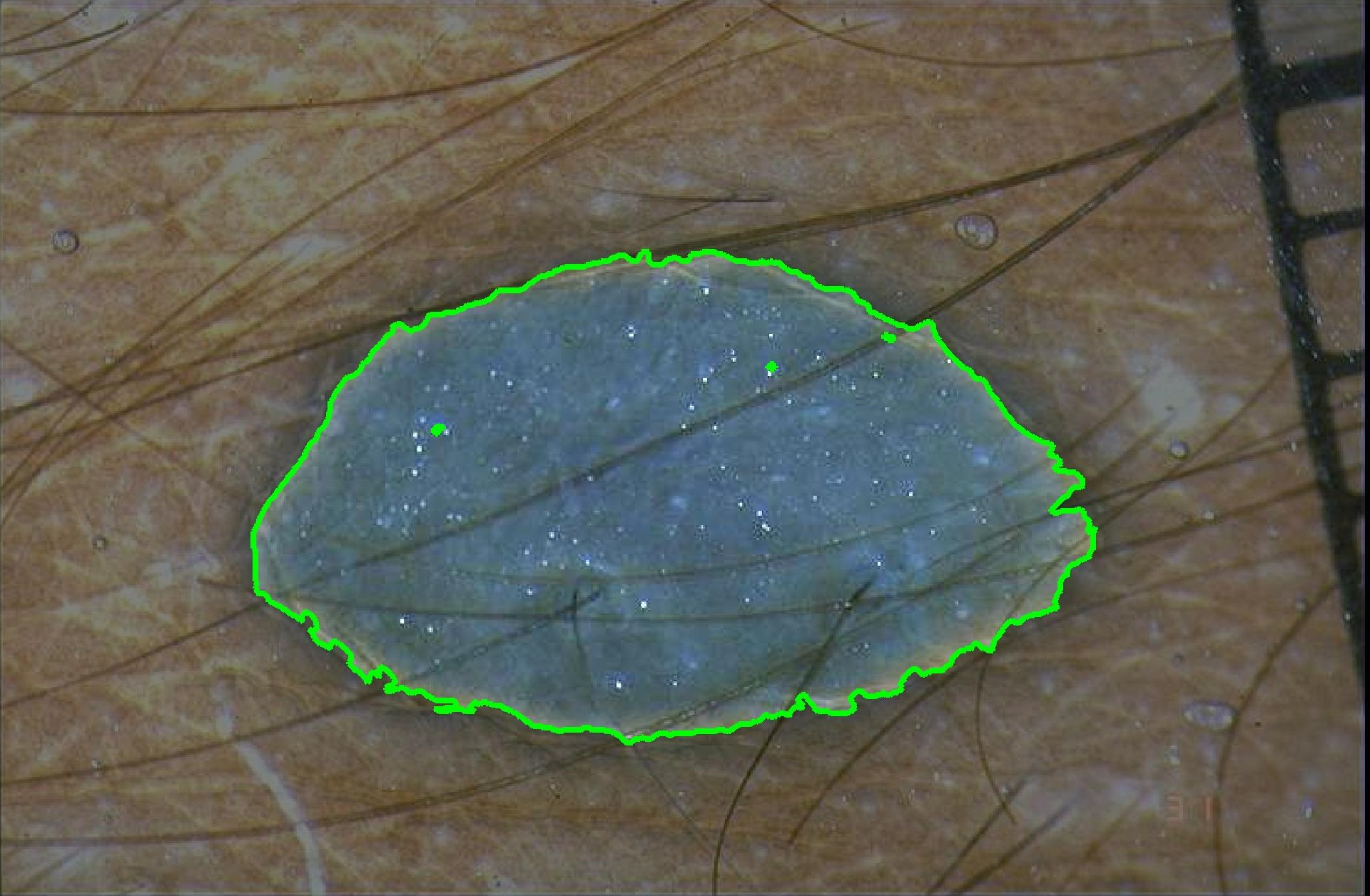} }
\\
\subfigure[]{\includegraphics[width=0.22\textwidth]{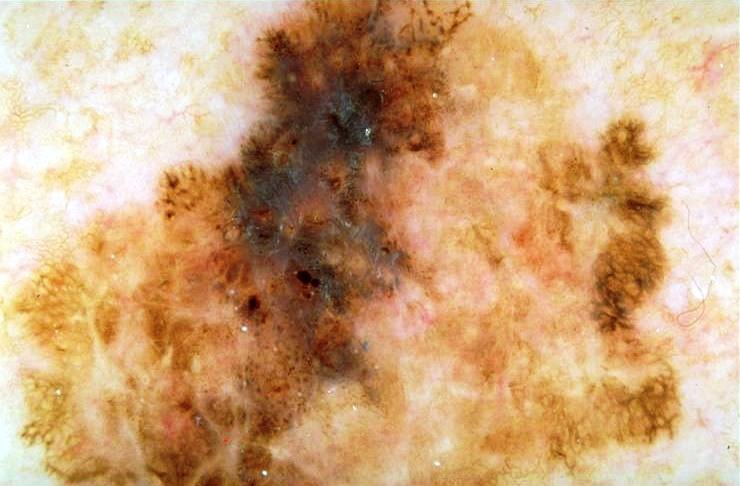} }
\subfigure[]{\includegraphics[width=0.22\textwidth]{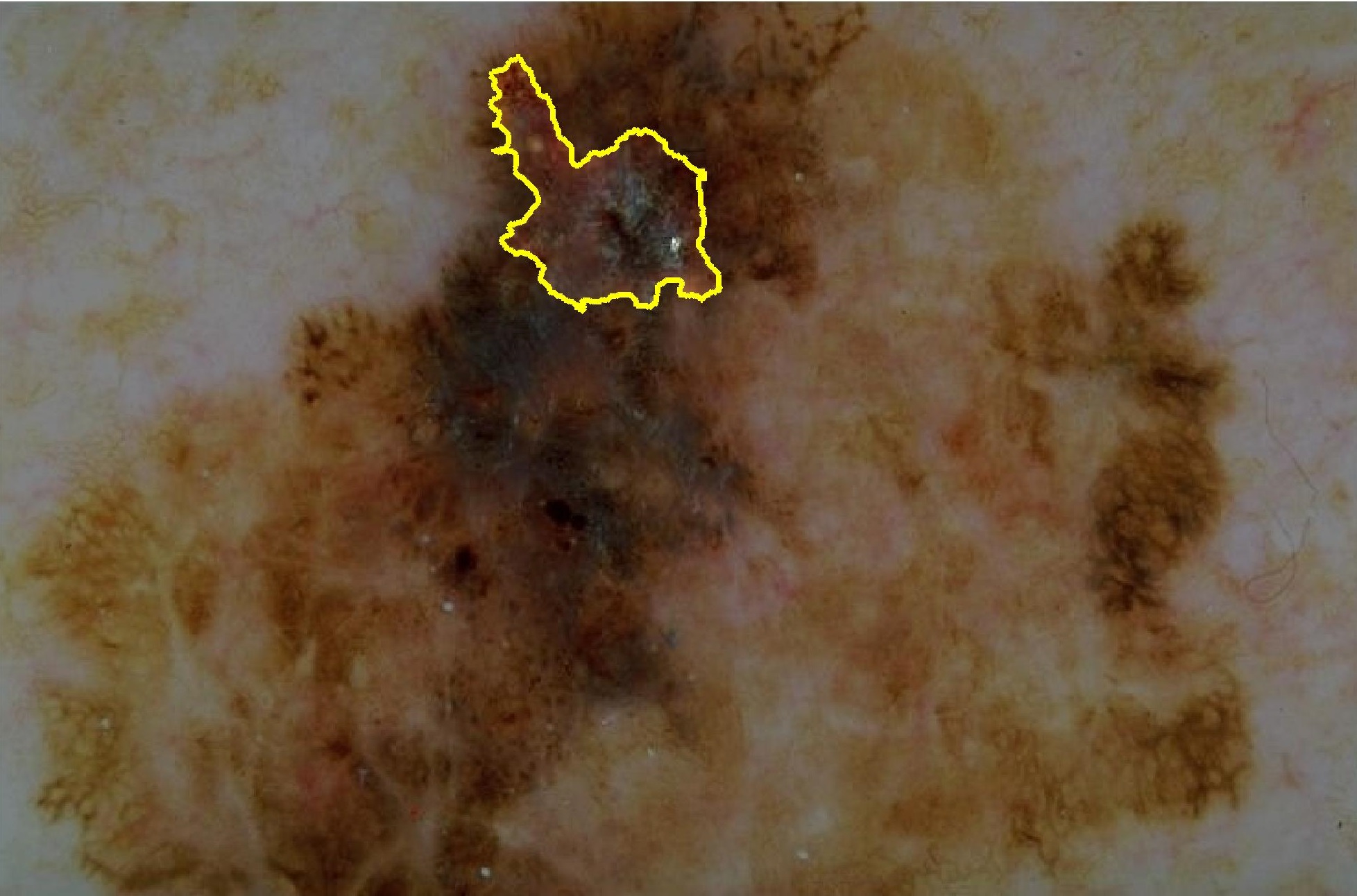} }
\subfigure[]{\includegraphics[width=0.22\textwidth]{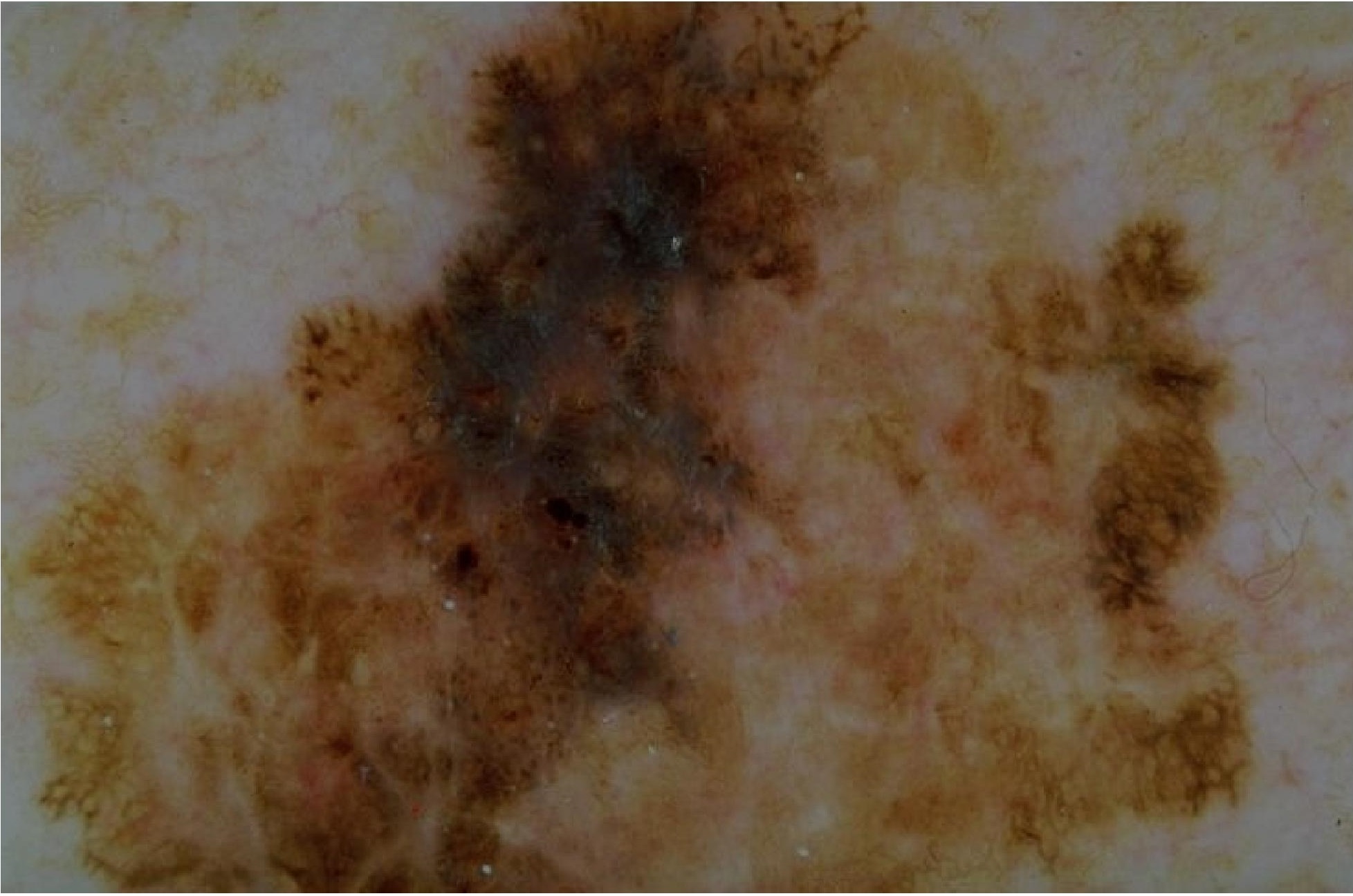} }
\subfigure[]{\includegraphics[width=0.22\textwidth]{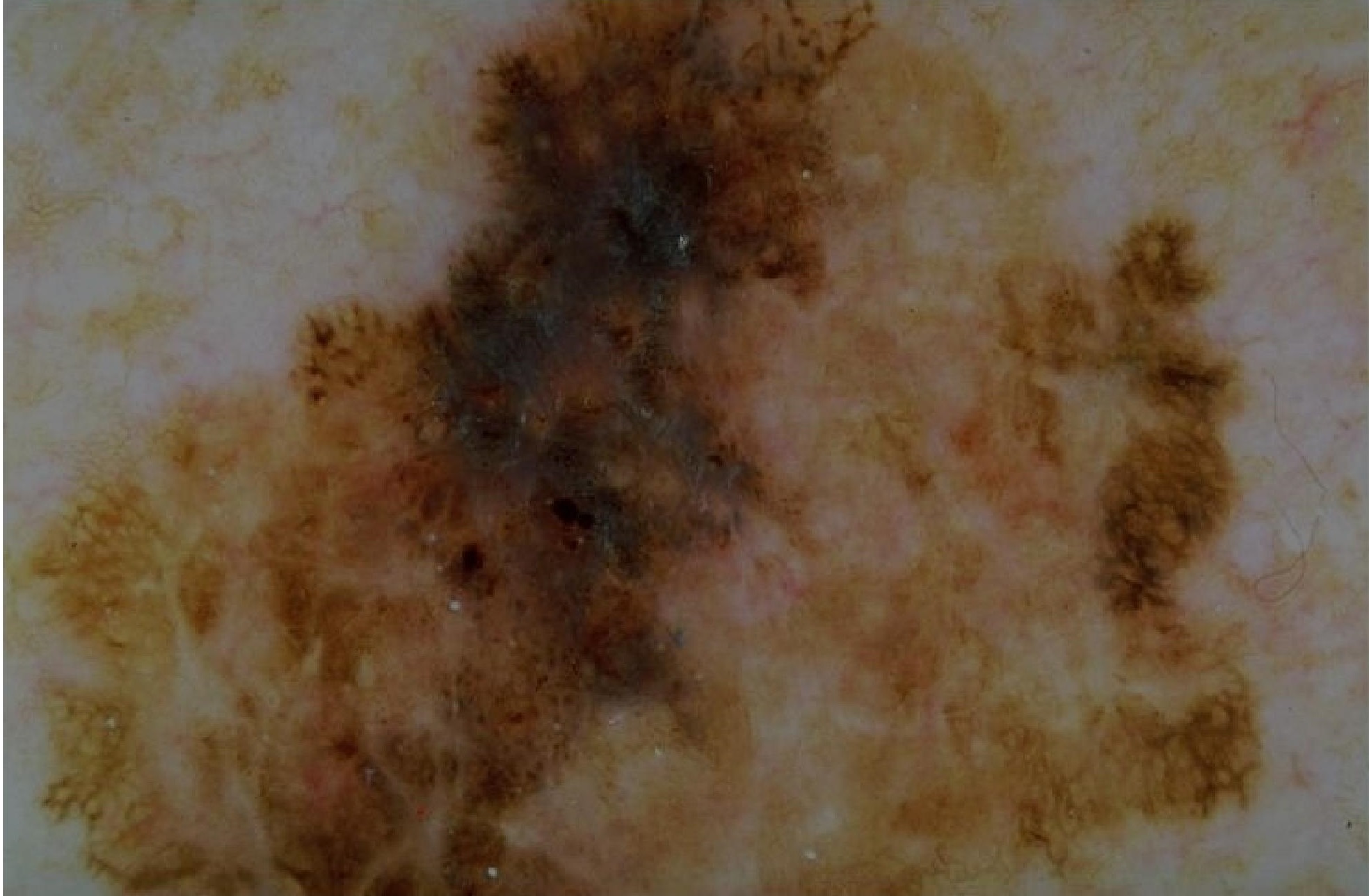} }
\\
\subfigure[]{\includegraphics[width=0.22\textwidth]{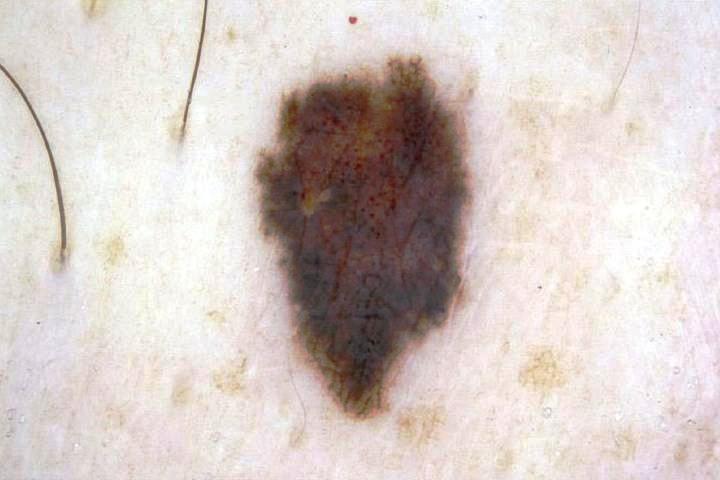} }
\subfigure[]{\includegraphics[width=0.22\textwidth]{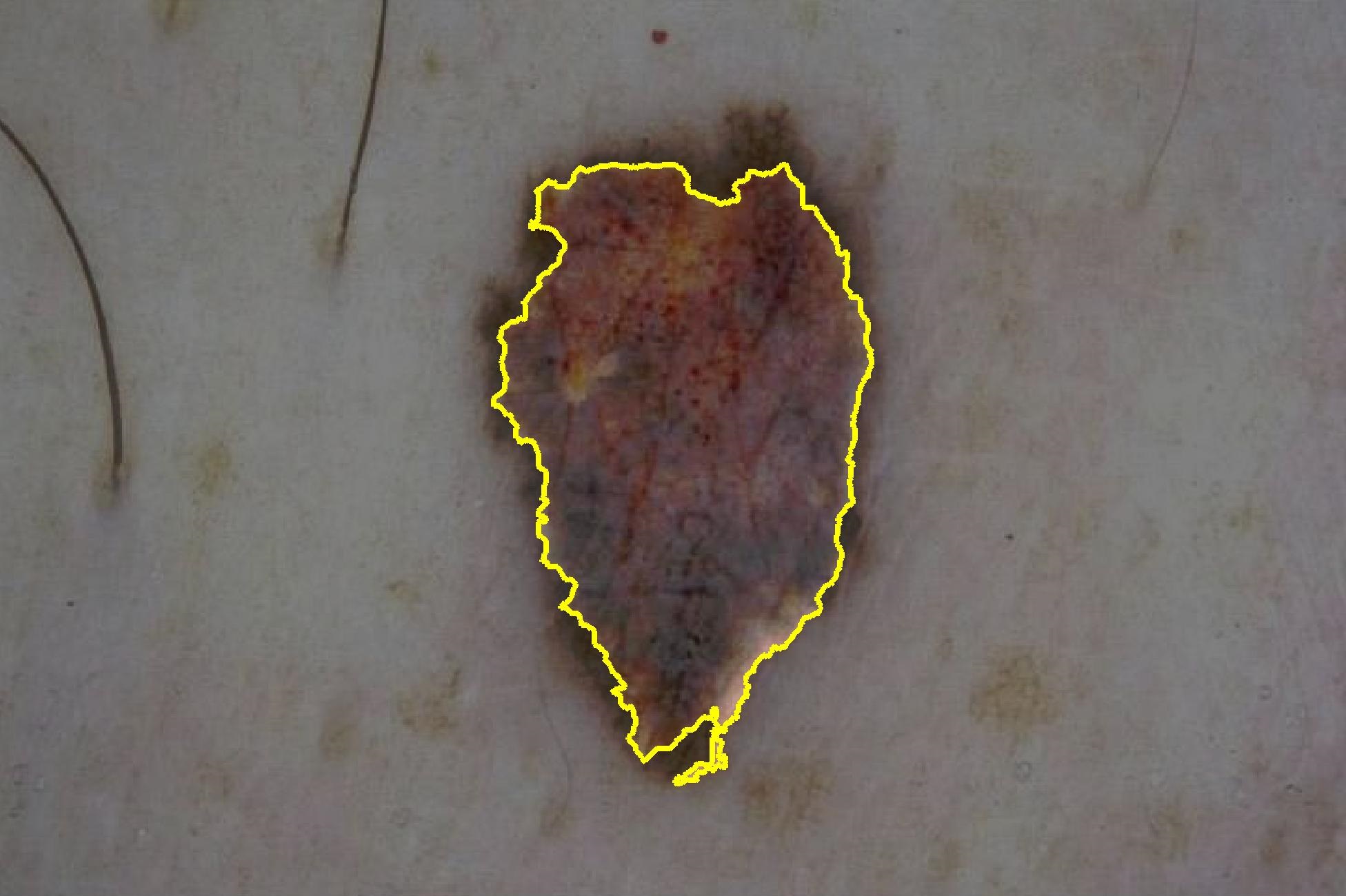} }
\subfigure[]{\includegraphics[width=0.22\textwidth]{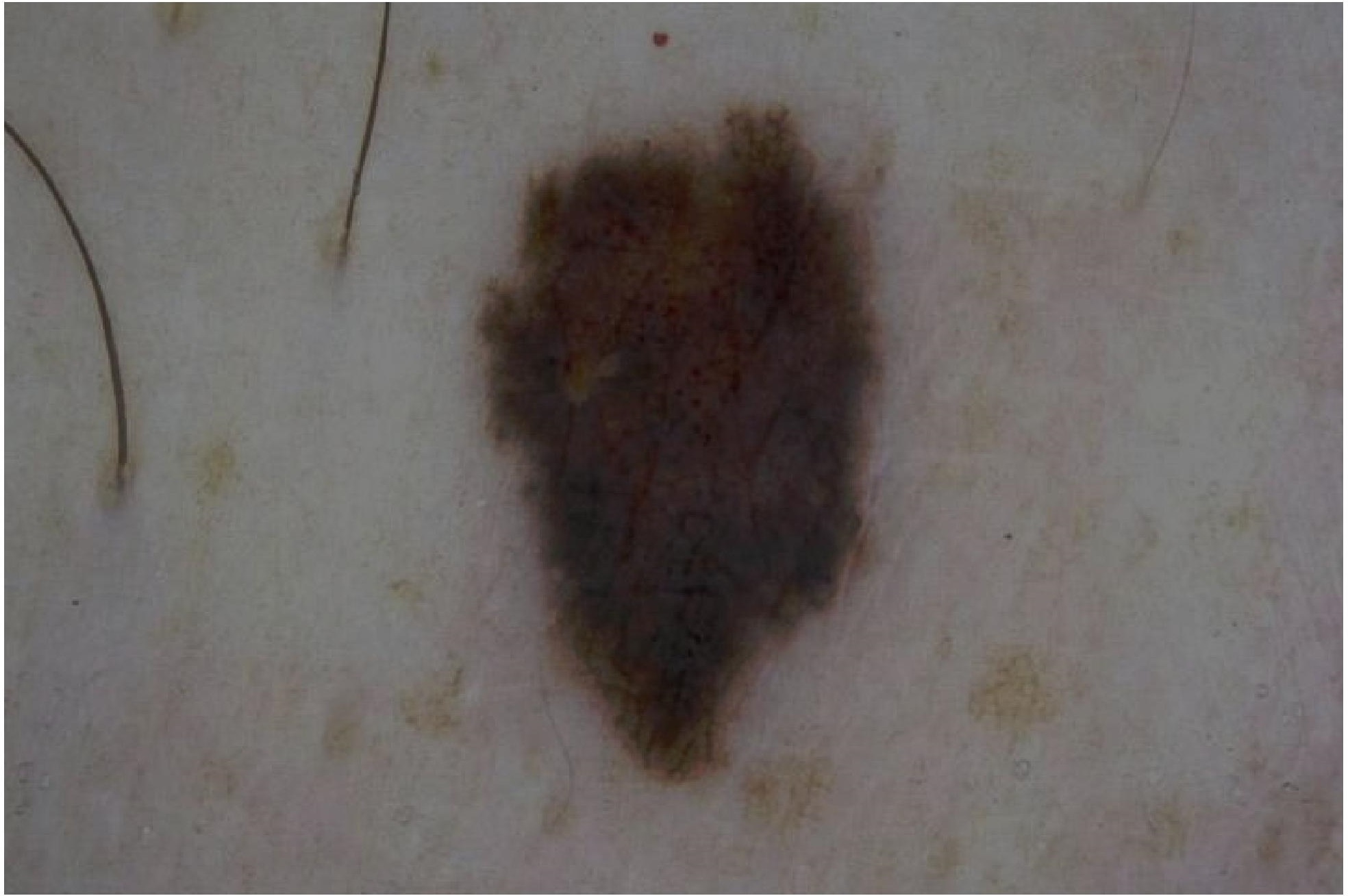} }
\subfigure[]{\includegraphics[width=0.22\textwidth]{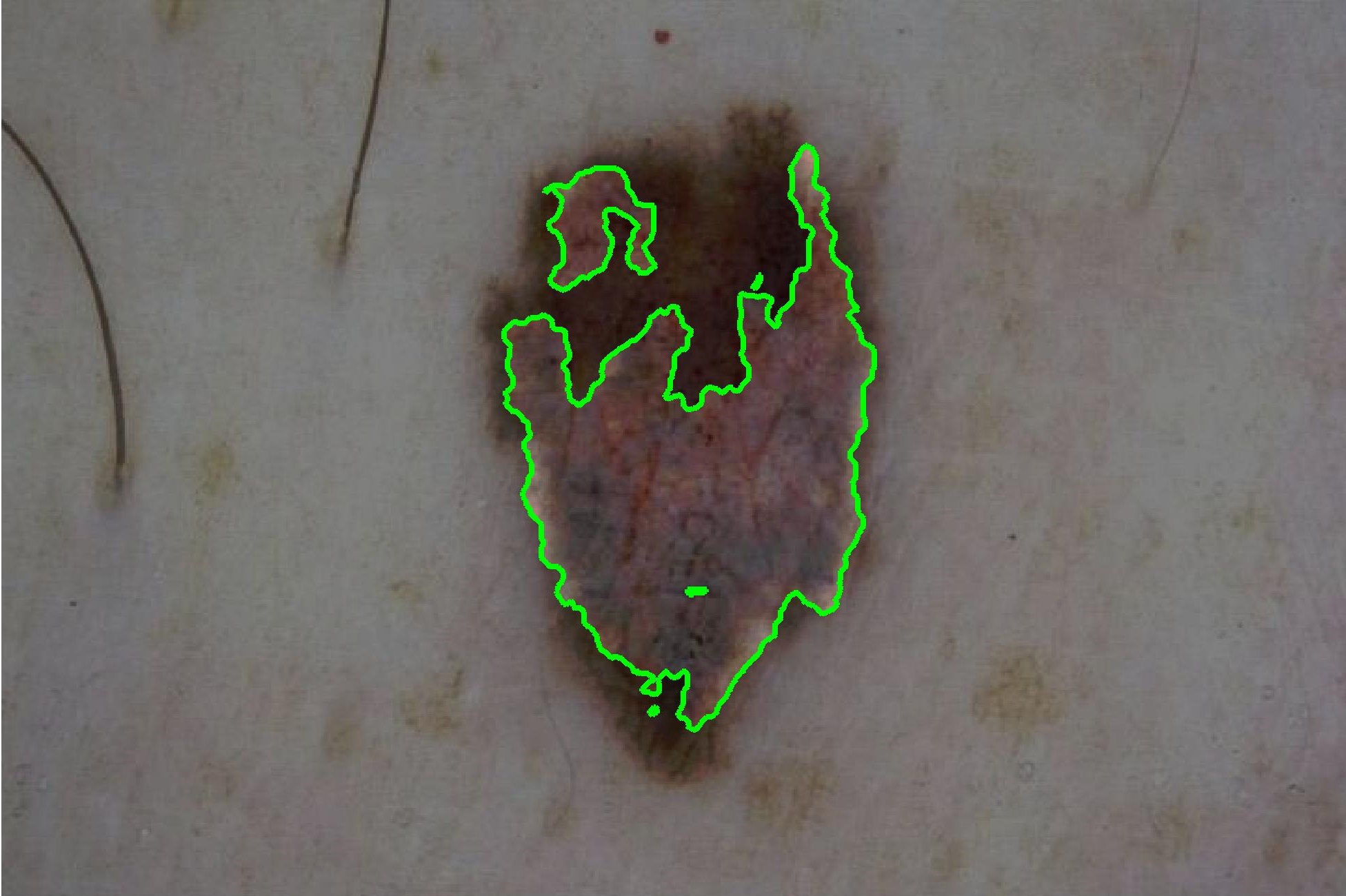} }

\caption{
Sample outputs: the first two rows are positive samples, with all detectors
succeeding in identifying and localizing the feature correctly. 
In the 2$^\text{nd}$ row, our method localized the feature over the areas that
contain both BWS and irregular globules. We believe that this supports our hypothesis that our proposed method learns to detect \textit{salient} regions for BWS identification.
This is further demonstrated in the 3$^\text{rd}$ row: both competing methods
detected confluent blue areas in a benign blue nevi falsely as BWS. Our method
correctly classified it as negative because there are no salient features
(blue/white colour plus other features such as globules) whereas
\cite{celebi_automatic_2008,madooei_automatic_2013} detected it incorrectly as
positive because they only look for colour features. Note that among all blue
and combined nevi, 43\% were falsely identified as positive by
\cite{celebi_automatic_2008}, 41\% by \cite{madooei_automatic_2013}, and 36\% by
our method, which further validates our approach.
The 4$^\text{th}$ row is a challenging positive example (small BWS area) which
only our proposed method succeeds in correctly identifying as positive. The
sample shown on the 5$^\text{th}$ row is an extremely challenging negative case. It contains blue-grey areas, atypical streaks, \textit{typical} pigment network and \textit{regular} globules. Our method (and \cite{madooei_automatic_2013}) wrongly detected this case as a positive. It was correctly labeled by \cite{celebi_automatic_2008} as a negative. For further examination, interested reader is referred to {\myverb} 
where we have provided sample outputs (all true-positive cases) of our proposed method.
}
\label{FIG:EX}
\end{figure*}

The methods of \cite{celebi_automatic_2008,madooei_automatic_2013} are simple to use and easy to understand, yet they impose disadvantages and shortcomings. For instance, their good sensitivity (recall) arrives at the expense of low specificity. Note that for propagating the label of pixels \cite{celebi_automatic_2008} and regions \cite{madooei_automatic_2013} to image-level, we applied a post processing step: an image is labelled positive if those pixels labelled positive were contained within the lesion. There are images containing bluish artefacts, e.g.\ ruler markings at the corner of Fig.\ref{FIG:EX}-i; the post-processing is set to reduce such false positives. 

Figure~\ref{FIG:EX} shows some sample outputs comparing the localization of BWS in test images among different methods. 
There are some interesting observations to be made here, particularly in support of our hypothesis that our proposed method learns to detect \textit{salient} regions for BWS identification. Please refer to the figure's caption for details. It is to be noted that the main limitation of the proposed method compared to \cite{celebi_automatic_2008,madooei_automatic_2013} (and any other supervised learning in general) is that wrong localization might still lead to correct image-level output. This is, however, a limitation of MIL in general and not specific to our case. 

Another general limitation of machine learning techniques, in particular supervised learning, is the issue of ``domain adaptation.''
The vast majority of learning methods today are trained and evaluated on the
same image distribution. A training dataset might be biased by the way in which
it was collected. A different dataset (visual domain) could differ by various
factors including scene illumination, camera characteristics, etc. Recent
studies (see e.g.\ \cite{torralba_unbiased_2011}) have demonstrated a
significant degradation in the performance of state-of-the-art image classifiers
due to domain shift. In dealing with this issue, a class of techniques, called
``Transfer Learning,'' has emerged that aims for developing domain adaptation
algorithms.\footnote{Transfer learning in general aims to transfer knowledge
between related domains. In computer vision, examples of transfer learning,
besides domain adaptation, include studies which try to ``overcome the deficit
of training samples for some categories by adapting classifiers trained for
other categories \cite{oquab_learning_2014}.'' A good review can be found in \cite{pan_survey_2010}.} 
Although Transfer Learning is beyond the scope of this study, we aim to examine and compare domain adaptability of our proposed method. To this aim, we tested our method and that of \cite{celebi_automatic_2008,madooei_automatic_2013} on a second database called  PH2 \cite{mendonca_ph2_2013}. 

The dermoscopic image database PH2 \cite{mendonca_ph2_2013} contains a total of
200 melanocytic lesions, including 80 common nevi, 80 atypical nevi, and 40
melanomas. This small database was built through a joint research collaboration
between the Universidade do Porto, T\'{e}cnico Lisboa, and the Dermatology service of Hospital Pedro Hispano in Matosinhos, Portugal. It includes clinical diagnosis and dermoscopic criteria, including presence (or absence) of blue-white veil and regression structures. 
Among the 200 images, 43 contain these features (positive cases) and the remaining 157 are free of these. This dataset is considerably less challenging compared to the Atlas, since most positive cases contain a sizeable BWS structure. 
Results are included in Table.\ref{tab:results} for comparison. Note that there is no training involved; all methods (both the proposed and baselines) were trained over the Atlas but tested on the PH2 set. The test results are consistent with our prior experiments over the Atlas. 

\section{Conclusion}
We proposed a new approach for automatic identification of the BWS feature which
needs considerably less supervision than for previous methods. 
Our method employs the MIL framework to learn from image-level labels, without explicit annotation of image regions containing the feature under study here. 
Experiments show that this method can learn, in addition to labelling the image,
to localize salient BWS regions in images, with high specificity, which
is of great importance in medical applications. Our results are very encouraging
since it is often the case that supervised learning with fully-labeled data
outperforms learning with only weakly-labeled data, with the performance of
latter being at best only comparable to that of the former. 
In future, we plan to adapt the multi-label multi-instance learning (MLMIL)
framework to simultaneously detect multiple dermoscopic features.



\clearpage
\bibliographystyle{IEEEtran}
\bibliography{TMI2015}

\end{document}